\documentclass[10pt,twocolumn,letterpaper]{article}

\usepackage[pagenumbers]{iccv} 

\definecolor{iccvblue}{rgb}{0.21,0.49,0.74}
\usepackage[pagebackref,breaklinks,colorlinks,allcolors=iccvblue]{hyperref}

\usepackage{amsmath}
\usepackage{graphicx}
\usepackage{algorithmic}
\usepackage[ruled,boxed]{algorithm2e}
\usepackage{setspace}
\usepackage{bbding}
\usepackage{booktabs}
\usepackage{makecell}
\usepackage{multirow}
\usepackage{bm}
\usepackage{caption}
\usepackage[capitalize]{cleveref}
\usepackage{cuted}
\usepackage{bbding}
\crefname{section}{Sec.}{Secs.}
\Crefname{section}{Section}{Sections}
\Crefname{table}{Table}{Tables}
\crefname{table}{Tab.}{Tabs.}
\crefname{algorithm}{Alg.}{Algs.}
\crefname{appendix}{App.}{Apps.}

\def\paperID{10466} 
\def\confName{ICCV}
\def\confYear{2025}

\title{Flexiffusion: Training-Free Segment-Wise Neural Architecture Search for Efficient Diffusion Models}

\author{Hongtao Huang\\
University of New South Wales\\
{\tt\small hongtao.huang@unsw.edu.au}
\and
Xiaojun Chang\\
University of Technology Sydney\\
{\tt\small XiaoJun.Chang@uts.edu.au}
\and
Lina Yao\\
CSIRO’s Data61 and University of New South Wales\\
{\tt\small lina.yao@data61.csiro.au}
}

\begin{document}
\maketitle
\begin{strip}
\begin{minipage}{\textwidth}\centering
\vspace{-25pt}
\includegraphics[width=0.99\textwidth]{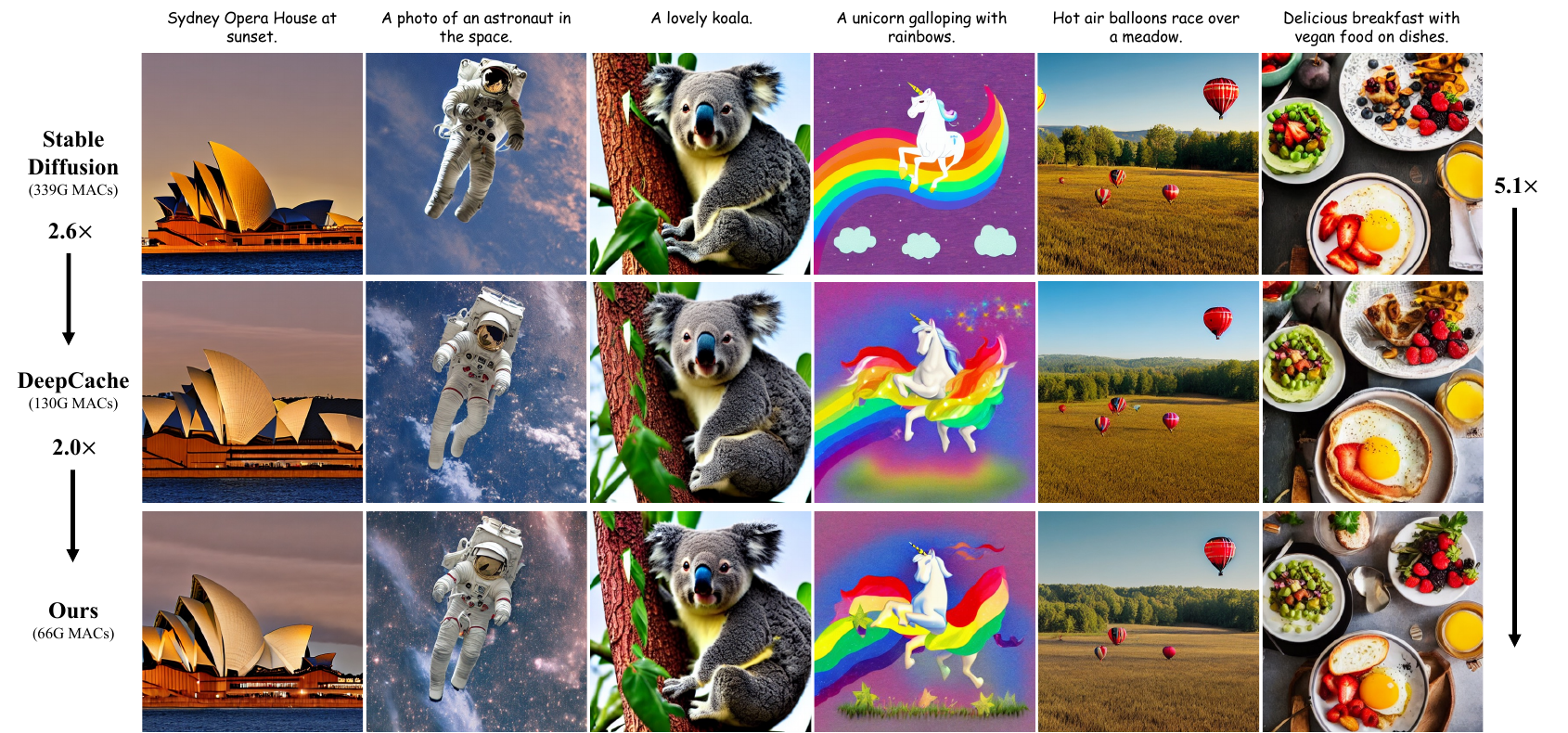}
\vspace{-5pt}
\captionof{figure}{Flexiffusion achieves a $5.1\times$ and $2.0\times$ reduction in computational cost compared to Stable Diffusion and DeepCache, respectively, while maintaining comparable image quality. The computational cost, shown as MACs, represents the average MACs for all steps.}
\label{figurelabel}
\end{minipage}
\end{strip}

\begin{abstract}
    Diffusion models (DMs) are powerful generative models capable of producing high-fidelity images but are constrained by high computational costs due to iterative multi-step inference. While Neural Architecture Search (NAS) can optimize DMs, existing methods are hindered by retraining requirements, exponential search complexity from step-wise optimization, and slow evaluation relying on massive image generation. To address these challenges, we propose Flexiffusion, a training-free NAS framework that jointly optimizes generation schedules and model architectures without modifying pre-trained parameters. Our key insight is to decompose the generation process into flexible segments of equal length, where each segment dynamically combines three step types: full (complete computation), partial (cache-reused computation), and null (skipped computation). This segment-wise search space reduces the candidate pool exponentially compared to step-wise NAS while preserving architectural diversity. Further, we introduce relative FID (rFID), a lightweight evaluation metric for NAS that measures divergence from a teacher model’s outputs instead of ground truth, slashing evaluation time by over $90\%$. In practice, Flexiffusion achieves at least $2\times$ acceleration across LDMs, Stable Diffusion, and DDPMs on ImageNet and MS-COCO, with FID degradation under $5\%$, outperforming prior NAS and caching methods. Notably, it attains $5.1\times$ speedup on Stable Diffusion with near-identical CLIP scores. Our work pioneers a resource-efficient paradigm for searching high-speed DMs without sacrificing quality.
\end{abstract}

\section{Introduction}
\label{sec:intro}

\begin{figure*}[t]
    \centering
    \includegraphics[width=0.9\linewidth]{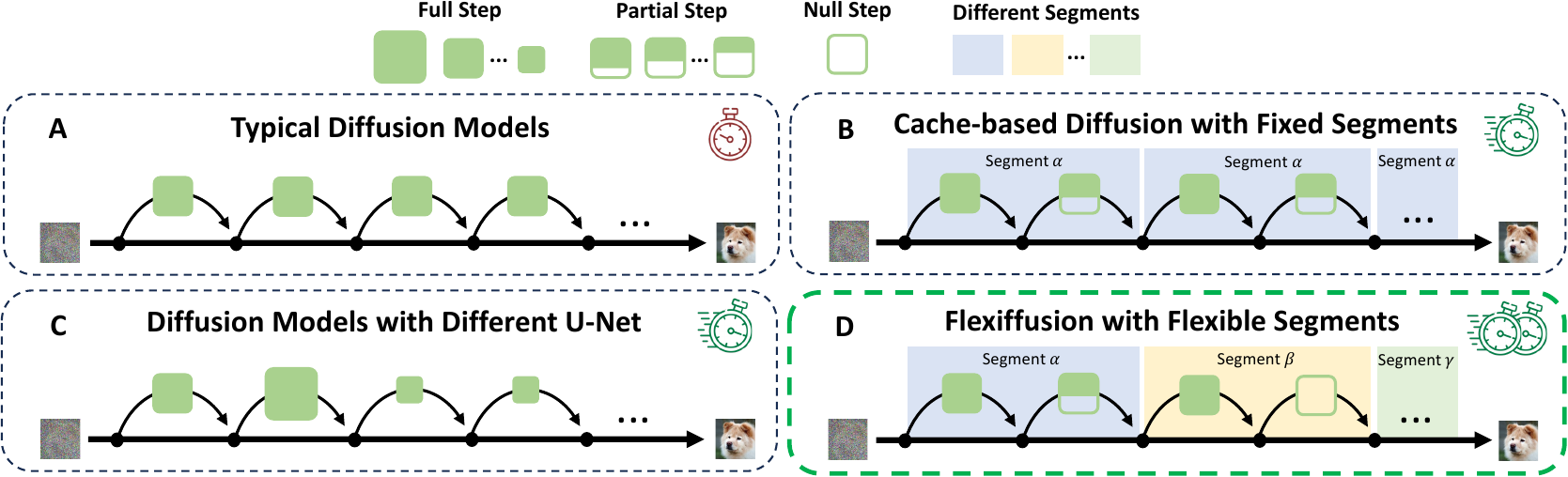}
    \caption{A comparison of different types of image generation schedules. \textbf{A)} Diffusion models with the same U-Net for each step; \textbf{B)} Diffusion models that can be divided into fixed segments via \textit{cache mechanism} \cite{ma2023deepcache}; \textbf{C)} Diffusion models with Different U-Nets for different steps; \textbf{D)} Flexiffusion with flexible segment settings further accelerates diffusion models by reducing generation redundancies.}
    \label{fig:flexiffusion}
\end{figure*}

Diffusion models (DMs) \cite{ho2020denoising,dhariwal2021diffusion,song2020denoising,choi2021ilvr,song2020score} ahave emerged as a powerful class of generative models, surpassing traditional approaches like variational autoencoder (VAE) \cite{kingma2013auto} and generative adversarial network (GAN) \cite{goodfellow2014generative}. At their core, DMs employ a U-Net architecture \cite{ronneberger2015u} to iteratively denoise images from Gaussian noise through a Markov chain process. While recent variants with transformer backbones \cite{peebles2023scalable} demonstrate potential for complex scene synthesis, U-Net-based DMs remain dominant in resource-constrained scenarios, such as mobile deployment \cite{li2024snapfusion}, due to their balance of efficiency and quality.

Despite their strengths, DMs suffer from a critical limitation: slow generation speeds caused by iterative multi-step inference. As shown in \cref{fig:flexiffusion} (A), a diffusion model is a combination of a denoising model (e.g., a U-Net) and a denoising schedule. 
The seminal DDPM \cite{ho2020denoising} models the generation as a Markov process, requiring a sampling schedule with 1000 steps. Current acceleration strategies fall into two categories. Schedule-based methods \cite{song2020denoising,liu2022pseudo,bao2022analytic,lu2022dpm} reduce step counts by reformulating the generation as non-Markov processes, but rely on heuristic step-selection rules that limit further optimization \cite{li2023autodiffusion}. Structure-based methods tackle per-step computation through techniques like pruning \cite{bolya2023tomesd,fang2024structural}, quantization \cite{he2024ptqd,shang2023post}, distillation \cite{salimans2022progressive,li2024snapfusion} or feature caching \cite{ma2023deepcache,wimbauer2024cache}. The latter cache-based methods, which reuse intermediate features across steps (\cref{fig:flexiffusion} (B)) offer training-free acceleration but enforce rigid caching policies across all steps. Crucially, most existing solutions either demand costly retraining or sacrifice flexibility by predefining acceleration rules, creating a tension between efficiency gains and model adaptability.

To overcome these trade-offs, Neural Architecture Search (NAS) \cite{zoph2016neural,real2019regularized,li2020block} has emerged as an automated paradigm for optimizing schedule and structural redundancies.  As illustrated in \cref{fig:flexiffusion} (C), NAS-based diffusion methods dynamically adapt network architectures across generation steps, for example, allocating lightweight subnets to less critical steps. However, existing NAS solutions introduce new bottlenecks: (1) they often require fine-tuning pre-trained models \cite{yang2023denoising,liu2023oms}, which is infeasible for large-scale DMs like Stable Diffusion \cite{rombach2022high} trained on 150K GPU hours; (2) The search space grows exponentially with step counts (e.g., $6^{100}$ candidates for 100 steps \cite{liu2023oms}); (3) Evaluation relies on generating over 10,000 images per candidate to compute metrics like FID \cite{heusel2017gans}, a process requiring days on modern GPUs. These limitations raise a pivotal question: \textit{How can we enable resource-efficient NAS for DMs that eliminates both schedule and structural redundancies without retraining or exhaustive evaluation?}

To address these challenges, we propose Flexiffusion, a training-free, segment-wise NAS framework that jointly optimizes sampling schedules and model architectures under resource constraints. Our approach operates on a pre-trained DM without modifying its parameter weights, and resolves the three bottlenecks by proposing a segment-wise search space and corresponding search method. Specifically, we introduce a segment-wise search space where the generation process is divided into equal-length segments. Each segment dynamically combines three step types: \textit{full}, \textit{partial}, and \textit{null}. The \textit{full} step execute the complete U-Net. The \textit{partial} step reuse cached features from preceding \textit{full} step to skip redundant computations. The \textit{null} entirely omit non-critical computations. Each segment starts with the \textit{full} step, followed by the \textit{partial} step and the \textit{null} step, as shown in \cref{fig:flexiffusion} (D). Consequently, the model search process is shifted from step-wise to segment-wise search, significantly reducing search space complexity and enabling efficient exploration within strict computational budgets.

To efficiently navigate the segment-wise search space, we design a lightweight evolutionary search algorithm tailored for diffusion models. Inspired by evolutionary NAS \cite{real2017large,real2019regularized}, the algorithm iteratively optimizes segment configurations through cycles of mutation and selection. Starting with a population of randomly initialized candidates, each iteration evaluates and selects top-performing models based on their speed-quality trade-offs, then generates new candidates by mutating their segment settings. Crucially, the evaluation overhead is minimized through our relative FID (rFID) metric, which replaces ground-truth comparisons with teacher-model alignment checks, reducing evaluation costs by an order of magnitude. This co-design of segment-aware search and efficient evaluation enables Flexiffusion to rapidly converge on high-quality architectures without retraining. Our main contributions are as follows:


\begin{itemize}
    \item We propose Flexiffusion, a NAS framework designed for diffusion acceleration by jointly optimizing denoising schedules and architectures without retraining.
    
    \item To overcome the combinatorial explosion in step-wise NAS, we introduce a segment-wise search space that divides the generation schedule into equal-length segments, significantly reducing the number of candidates while preserving diversity. Based on this space, we propose a corresponding segment-wise search algorithm.
    
    \item To reduce the NAS evaluation cost, we design a faster metric for evaluation in NAS, termed relative FID, which measures alignment between generated images and those from the original diffusion model (teacher), rather than ground-truth data. rFID achieves 10× faster evaluation while maintaining high ranking consistency.

    \item Flexiffusion seamlessly integrates with existing samplers (DDIM \cite{song2020denoising}, PLMS \cite{liu2022pseudo}, DPM-Solver \cite{lu2022dpm}) and frameworks (DDPM \cite{ho2020denoising}, LDM \cite{rombach2022high}, Stable Diffusion \cite{rombach2022high}). Extensive experiments show that the diffusion models searched by Flexiffusion achieve an improved balance between image quality and generation speed, with exceptional performance in lightweight model scenarios.

\end{itemize}
\section{Related Work}
\subsection{Diffusion Models and Efficient Sampling}
\label{subsec: Diffusion Models and Efficient Sampling}

Diffusion models \cite{ho2020denoising,song2020denoising,rombach2022high} are the cutting-edge generative models that generate diverse images from the Gaussian distribution. Despite their significant advancements, diffusion models suffer from inherent low generation speed due to their step-by-step sampling processes \cite{song2019generative,song2020denoising}. Current efficient sampling strategies can be categorized into schedule-based and structure-based acceleration.

\noindent\textbf{Schedule-based acceleration.} It focuses on reducing the number of steps in the generation schedule. Since the Markov-based generation model DDPM \cite{ho2020denoising}, many pioneering works have applied numerical analysis as faster samplers \cite{song2020denoising,song2020score,lu2022dpm,liu2022pseudo}. DDIM \cite{song2020denoising} reduces the sampling steps in the schedule by recomposing a non-Markovian process. Some approaches reformulate the generation process using SDEs or ODEs \cite{bao2022analytic,lu2022dpm} to further streamline the schedule, reducing it to only 10-25 model inference times. These mathematical methods are training-free, allowing the pre-trained U-Net model to be reused with a proper sampler \cite{song2020score}. To further reduce the steps, some training-based methods replace the remaining steps with a VAE \cite{lyu2022accelerating} or distill a new model with fewer steps \cite{salimans2022progressive}. The Consistency Model \cite{song2023consistencymodels} uses a consistency formulation enabling single-step generation, but it requires training from scratch or distillation from a pre-trained model.

\noindent\textbf{Structure-based acceleration.} It prefers to reduce the computational cost of U-Net at each step. A range of lightweight model techniques, such as structure pruning \cite{fang2024structural,li2024snapfusion}, knowledge distillation \cite{luhman2021knowledge,kim2023bksdm}, and model quantization \cite{he2024ptqd,shang2023post}, have been applied to diffusion models to enable faster inference. While these approaches use the same model throughout, multi-model methods \cite{liu2023oms,yang2023denoising} applied a set of different U-Net models in different steps. There are also plug-and-use modules for structure lightweight such as ToMe \cite{bolya2023tomesd}. Additionally, cache-based diffusion models propose the \textit{cache mechanism} \cite{ma2023deepcache,wimbauer2024cache} that stores feature maps from the current step for reusing, omitting the computing of parts of U-Net blocks in the subsequent steps.

\subsection{Neural Architecture Search}
Neural Architecture Search (NAS) is a pivotal subfield of Automated Machine Learning (AutoML) \cite{he2021automl}, dedicated to discovering optimal neural architectures under diverse resource constraints \cite{zoph2016neural,real2019regularized,zoph2018learning,ren2021comprehensive}. The NAS workflow comprises two phases: (1) constructing a search space encompassing architectures with varying hyperparameters; (2) employing search algorithms to identify high-performing candidates through iterative training and evaluation.

A critical challenge in NAS lies in balancing search diversity against evaluation efficiency, as exhaustively training each candidate from scratch incurs prohibitive computational costs. To mitigate this, block-wise NAS \cite{li2020block,li2021bossnas} introduces a modular paradigm where architectures are decomposed into reusable blocks. By limiting searches to block configurations rather than step configurations, these methods drastically shrink the candidate pool while preserving design flexibility. 

\subsection{NAS for Efficient Diffusion Models}
\label{subsec:diffusion models with nas}
Recent works \cite{yang2023denoising,liu2023oms,li2023autodiffusion} have explored integrating Neural Architecture Search (NAS) with diffusion models to automate the design of efficient schedules and architectures. AutoDiffusion \cite{li2023autodiffusion} suggests a non-uniform skipping of steps and network structural blocks. OMS-DPM \cite{liu2023oms} first trained several candidate models for individual denoising steps. DDSM \cite{yang2023denoising} further extends supernet-based NAS supernet-based NAS \cite{yu2019autoslim} to dynamically sample lightweight U-Net variants across steps. Despite their promise, current methods face critical limitations that hinder practical adoption.

\vspace{1pt}
\noindent\textbf{Extra training cost.} Many NAS-based diffusion methods necessitate extra retraining or fine-tuning for the pre-trained U-Net. However, contemporary high-performing diffusion models are characterized by their substantial size, complexity, and demanding training requirements, necessitating vast amounts of training data and intricate training processes. For example, the training cost of Stable Diffusion \cite{rombach2022high} is about 150000 GPU hours on 256 NVIDIA A100 GPUs. Retraining such a model impose prohibitive training costs. Therefore, our method prefers training-free NAS, which searches based on pre-trained high-performing models.

\vspace{1pt}
\noindent\textbf{Explosive number of candidates.} Current NAS-based diffusion methods are typically step-wise searching. However, a diffusion model involves a considerable number of inference steps (e.g., 100 steps in DDIM \cite{song2020denoising}) and the search space complexity explodes exponentially with step counts. Assuming there are six U-Net models for each step as in OMS-DPM \cite{liu2023oms}, the total number of candidate diffusion models could reach up to $6^{100}$, rendering effective exploration infeasible. Motivated by block-wise NAS \cite{li2020block,li2021bossnas}, we propose a segment-wise search space that groups steps into reusable segments, significantly reducing complexity.

\vspace{1pt}
\noindent\textbf{Time-consuming model evaluation.} NAS relies on performance evaluation for each candidate to identify the optimal model. However, the computationally intensive nature of diffusion models poses a significant challenge.  Additionally,  FID \cite{heusel2017gans}, the commonly used metric, requires over 10,000 images for accurate evaluation, further slowing down evaluation and limiting thorough exploration of the search space. To address this, we propose a faster metric, relative FID (rFID), which replaces ground-truth comparisons with alignment checks between images generated by the teacher model and those by the student models.

\section{Preliminary} 
\label{sec: preliminary}
\begin{figure}[t]
\centering
\includegraphics[width=0.90\linewidth]{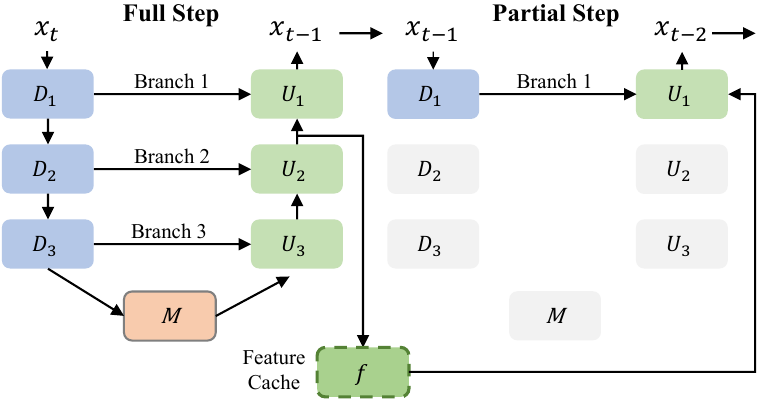}
\vspace{-5pt}
\caption{The structure of U-Net and an example of a segment with cache. The \textit{full} step provides a cache for the \textit{partial} step.}
\label{fig:unet_and_cache}
\end{figure}

\textbf{U-Net.} Many diffusion models \cite{ho2020denoising,song2020denoising,rombach2022high} use the U-Net \cite{ronneberger2015u} as a backbone architecture for image generation. U-Net consists of an equal number of downsampling and upsampling blocks, with each pair of symmetrical blocks connected by a skip connection. Here we use $D_b$ to represent the $b$-$th$ downsampling block and $U_b$ is the corresponding upsampling block, as shown in \cref{fig:unet_and_cache}. The data flow has multiple traversing paths: a block-by-block mainstream and those skipping branches. Assuming there are total $B$ skipping branches in the U-Net and $T$ steps in the sampling schedule, we formulate the concatenation operation set $\mathcal{C}_b$ at the $b$-th branch among the whole schedule as below:
\vspace{-3pt}
\begin{equation}\label{equ: u-net}
    \mathcal{C}_b\!=\!\bigg\{\!\!\left\{d_b^{(T)}\!\!\oplus\!u_{b+1}^{(T)}\!\right\}\!\cdots\!\left\{d_b^{(2)}\!\!\oplus\!u_{b+1}^{(2)}\!\right\}\!\!,\left\{d_b^{(1)}\!\!\oplus\!u_{b+1}^{(1)}\!\right\}\!\!\bigg\},
\end{equation}
where $d_b^{(T)}$ and $u_{b+1}^{(T)}$ is the output feature maps from $D_b$ and $U_{b+1}$ at time step $T$, respectively. $\oplus$ denotes the feature concatenation operation. The concatenation operation set for all branches can be represented as $\mathcal{C}=\{\mathcal{C}_1, \mathcal{C}_2\cdots\mathcal{C}_B\}$. The primary bottleneck in inference speed arises from the large number of $B$ and $T$.

\begin{figure}[t]
\centering
\includegraphics[width=0.99\linewidth]{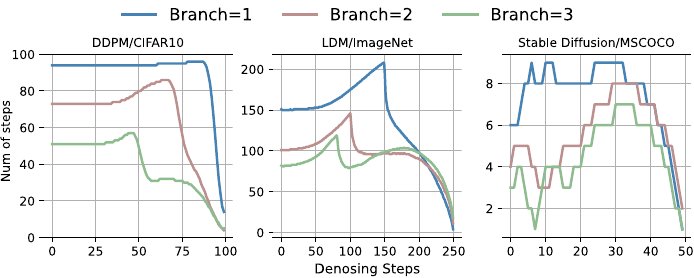}
\vspace{-5pt}
\caption{The number of steps where the cosine similarity between the feature maps and the current step’s feature maps exceeds 0.9. Those feature maps are from three different branches, respectively.}
\label{fig:similarity}
\end{figure}

\begin{figure*}[t]
\centering
\includegraphics[width=0.95\linewidth]{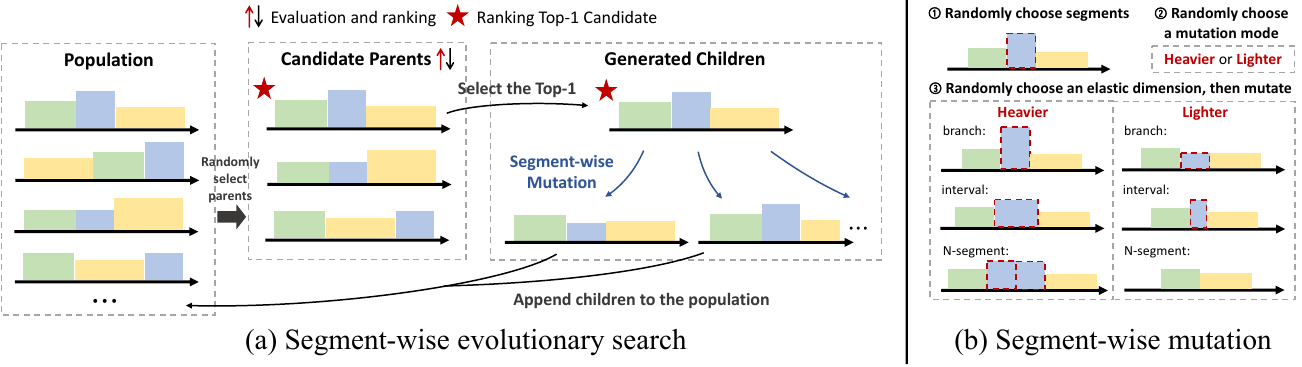}
\vspace{-5pt}
\caption{\textbf{(a):} A workflow of the segment-wise search. The computational cost primarily arises from the ranking process; \textbf{(b):} The segment-wise mutation process. The heavier mode increases computation by caching deeper U-Net branches, replacing \textit{null} steps with \textit{partial} steps, or splitting segments to increase \textit{N-segment}. The lighter mode reduces computation by using shallower cached branches, replacing \textit{partial} steps with null steps, or merging segments to decrease \textit{N-segment}.}
\label{fig:evolutionary search and mutation}
\end{figure*}

\vspace{1pt}
\noindent\textbf{Cache Mechanism.} Latest approaches for diffusion acceleration, CacheMe \cite{wimbauer2024cache} and DeepCache \cite{ma2023deepcache}, speed up the image generation by reusing portions of U-Net features for subsequent steps, based on the observation that these features show significant similarity across adjacent steps. In the following of this paper, the \textit{cache mechanism} is based on DeepCache, as it does not require extra model training. 

Assuming the generation schedule is uniformly divided into equal-length segments, with each segment containing $n$ steps. For a segment $\{t_{n} \cdots t_2, t_1\}$, the output feature maps $u^{(t_n)}_{b+1}$ of the $(i\!\!+\!\!1)$-$th$ upper block at the first step $t_n$ are stored as a cache $f$ and reused in the following steps. The concatenation operation applied at the $b$-th branch within this segment is formulated as below:
\vspace{-2pt}
\begin{align}
    \mathcal{C}_{b}^{(t_{n}\rightarrow t_1\!)}\!=\!\bigg\{\!\!\left\{d_b^{(t_n)}\!\oplus\!f\right\}\!\cdots\!\left\{d_b^{(t_2)}\!\oplus\!f\right\}\!,\left\{d_b^{(t_1)}\!\oplus\!f\right\}\!\!\bigg\}\label{equ:cache}\!,\\
    \text{where}\  f=u^{(t_n)}_{b+1} \label{equ:cache_where}
\end{align}

\vspace{-2pt}
As shown in \cref{fig:unet_and_cache}, the first step within a segment, which provides feature cache $f$, is the \textbf{\textit{full}} step, while the step using the cache $f$, is the \textbf{\textit{partial}} step. Compared to a typical diffusion model that only consists of \textit{full} step as shown in \cref{fig:flexiffusion} (A), the cache-based one uses \textit{partial} steps for acceleration. The more \textit{partial} steps involved, the greater the speed-up, but with a decrease in image quality. As for a segment with $n$ steps, DeepCache sets a $full$ step with $n\!-\!1$ \textit{partial} steps, and all segments are set the same $n$ and $b$.  In this paper, we introduce another step, the \textbf{\textit{null}} step, which skips all the computation, to reduce the redundancy in generation process.

The success of the \textit{cache mechanism} relies on the similarity of feature maps in adjacent steps.  \cref{fig:similarity} reports the number of steps where the cosine similarity between their features and the current step's features is higher than $90\%$. For instance, in the middle figure of \cref{fig:similarity}, the $branch=1$ feature maps at the $150$ step report over 200 similar steps. 

Although there are strong similarities in the schedule, the ratio of similar to total steps varies across different frameworks, datasets, and branch settings. Therefore, rather than fixed cache settings for all segments in DeepCache, we search for flexible cache settings that can fully utilize those similarities and explore potential models.





\vspace{1pt}
\noindent\textbf{Varying Importance of Different Steps.} Diffusion models have a step-by-step generation process while each step within the schedule exhibits different behaviors \cite{choi2022perception,deja2022analyzing}. \cref{fig:similarity} not only shows the similarity in adjacent steps but also demonstrates that different steps have different roles in the schedule. Steps toward the end of the schedule exhibit lower similarity to other steps, showing their importance. In this case, our target is to identify those steps with higher importance and omit the lower ones via NAS. 



\section{Segment-Wise Neural Architecture Search}
\label{sec:segment-wise NAS}


We present Flexiffusion, a segment-wise neural architecture search (NAS) framework designed to accelerate diffusion models through joint optimization of sampling schedules and architectural configurations. Motivated by two core insights, our approach builds upon the \textit{cache mechanism}: (1) Its inherent feature reuse capability eliminates the need for U-Net retraining or fine-tuning; (2) Its native grouping of sequential steps into functional units aligns perfectly with our segment-wise search paradigm. Specifically, we decompose the generation process into equal-length segments where each segment is configured with full (complete computation), partial (cache-reused), and null (skipped) steps through learnable configurations. The remainder of this section details our three key innovations: a segment-wise search space, a segment-wise evolutionary search strategy and a resource-efficient performance estimation metric.

\subsection{Segment-Wise Search Space}
\label{subsec:segment-wise search space}

To create a search space with diverse candidate variants based on the pre-trained diffusion models, we extends the \textit{cache mechanism} by introducing three search dimensions: 
\begin{itemize}
    \item \textbf{\textit{N-segment}} controls the total number of segments in the generation schedule, enabling coarse-grained control over computational budgets.
    \item \textbf{\textit{branch}} determines which block’s features are cached, balancing feature reuse and reconstruction quality.
    \item \textbf{\textit{interval}} specifies the number of active steps (\textit{full}/\textit{partial}) before skipping (\textit{null} steps). For example, a segment configured as \textit{full} → \textit{partial} → \textit{null} → \textit{null}, the \textit{interval} has an interval of 4.
\end{itemize}

Unlike the vanilla caching approach, which rigidly alternates full and partial steps for all the segments, our search space allows candidates to have dynamic and flexible step-type ratios for each segment. This design maximizes redundancy removal while retaining essential computations.

Thus, a diverse search space is constructed based on the varying segment configurations. Assuming that there are three different choices for each dimension: \textit{N-segment} choices $\{s_1, s_2, s_3\}$, \textit{branch} choices $\{b_1, b_2, b_3\}$ and interval choices $\{n_1, n_2, n_3\}$. The length of the generation schedule is $s_3\times n_3$. The total number of candidates in the search space is equal to $(3\times3)^{s_1}+(3\times3)^{s_2}+(3\times3)^{s_3}$. The complexity for our segment-wise search space is $\mathcal{O}(9^{s_3})$, while step-wise search space \cite{liu2023oms,yang2023denoising} will exponentially increases to $\mathcal{O}(9^{s_3\times n_3})$. We provides detailed examples of searched candidates in the supplementary material.

\subsection{Segment-Wise Evolutionary Search}
\label{subsec: evolutionary search}
Based on evolutionary NAS\cite{real2017large}, we propose our segment-wise evolutionary search algorithm efficiently explore our search space as shown in \cref{fig:evolutionary search and mutation} (a). Our segment-wise search operates as an iterative loop that progressively refines candidate models until a predefined termination condition (e.g., maximum iterations) is met. The algorithm iteratively evolves a population of candidates through three phases:
\begin{itemize}
    \item \textbf{Population Initialization:} Generate an initial population by randomly sampling candidates from our segment-adaptive search space (\cref{subsec:segment-wise search space}), with candidate configurations defined by \textit{N-segment}, \textit{branch}, and \textit{interval}.
    \item \textbf{Evaluation \& Selection:} Rank candidates using our designed relative FID (rFID) metric (\cref{subsec:performance estimaton}) and top performers are selected as parents for mutation.
    \item \textbf{Segment-Aware Mutation:} Randomly apply heavier or lighter mutations to parents as shown in \cref{fig:evolutionary search and mutation} (b).
\end{itemize}

Those mutated candidates are added back to the population, and the second and third phases are iterated until the predefined maximum number of iterations is reached. This search process gradually converges toward architectures that maximize the speed-quality trade-offs, with the highest-ranked candidate identified as the optimal solution. Therefore, the algorithm respects the segment hierarchy by performing mutations on entire segments rather than individual steps, ensuring both search efficiency and architectural coherence. Pseudo-code of the search and mutation process are provided in the supplementary material.

\begin{figure}[t]
\centering
\includegraphics[width=0.88\linewidth]{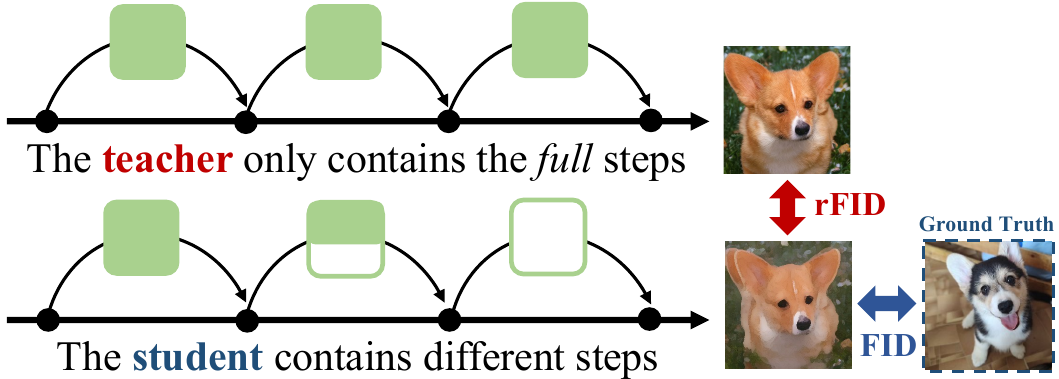}
\vspace{-5pt}
\caption{The difference between rFID and FID.}
\label{fig:relative fid}
\end{figure}

\subsection{Efficient Performance Estimation}
\label{subsec:performance estimaton}
A major bottleneck in NAS is the efficient evaluation of candidate models. Traditional metrics like FID \cite{heusel2017gans} require generating and comparing over 10,000 images, a process that can consume hundreds of GPU hours. While reducing the image count to 1,000 accelerates evaluation, it severely compromises ranking accuracy.

To resolve this, we propose relative FID (rFID), a lightweight metric inspired by knowledge distillation \cite{hinton2015distilling}. Unlike FID, which measures divergence from ground-truth data, rFID evaluates candidates by their alignment with a teacher model, the original diffusion model using full steps. By feeding identical input noise to both teacher and candidate models, we ensure semantic consistency between their outputs enabling reliable quality assessment with only 1,000 images. \cref{fig:relative fid} illustrated the difference between FID and rFID. Besides, we ablate alternative metrics like pixel-wise Mean Squared Error (MSE), which fails to capture perceptual quality and favors blurry outputs over realistic ones. rFID achieves faster evaluation than FID while maintaining high ranking accuracy, making it ideal for training-free NAS. A detailed quantitative comparison of FID, rFID, and MSE is presented in \cref{subsubsec: study for rfid}.

\begin{table*}[t]
    \centering
    \resizebox{0.98\linewidth}{!}{
    \begin{tabular}{c||ccccc|cccc}
    \toprule
    \multicolumn{10}{c}{\textbf{ImageNet} \bm{$256\times256$}} \\
    \midrule
    \textbf{Method} & \textbf{NFE} $\downarrow$ & \textbf{s/image} $\downarrow$ & \textbf{avg. MACs} $\downarrow$ & \textbf{Speedup} $\uparrow$ & \textbf{Retrain} &\textbf{FID} $\downarrow$ & \textbf{IS} $\uparrow$ & \textbf{Precision} $\uparrow$ & \textbf{Recall} $\uparrow$ \\
    \midrule
    LDM        & 250  & 5.08 & 99.82 G  & $1.0\times$ & \XSolidBrush & 3.37 & 204.56 & 82.71 & 53.86 \\
    DDSM*    & 250 & 3.22 & 19.40 G  & $5.1\times$ & \Checkmark   & 24.71  & - & - & - \\
    \midrule
    DeepCache-A & 250 & 2.68 & 52.12 G & $1.9\times$ & \XSolidBrush & 3.39 & 204.09 & 82.75 & 54.07 \\
    DeepCache-B & 250 & 1.26 & 23.50 G & $4.2\times$ & \XSolidBrush & 3.55 & 200.45 & 82.36 & 53.30 \\
    DeepCache-C+ & 250 & 0.75 & 13.97 G & $7.1\times$ & \XSolidBrush & 4.27 & 193.11 & 81.75 & 51.84 \\
    DeepCache-D+ & 250 & 0.55 & 9.39 G & $10.6\times$ & \XSolidBrush & 7.11 & 167.85 & 77.44 & 50.08 \\
    \midrule
    \textbf{Flexiffusion-A} & 174 & 2.08 & 39.26 G & $2.4\times$ & \XSolidBrush & 3.37 & 203.10 & 82.87 & 53.98 \\
    \textbf{Flexiffusion-B} & 69  & 0.86 & 15.67 G & $6.4\times$ & \XSolidBrush & 3.68 & 198.95 & 82.36 & 52.99 \\
    \textbf{Flexiffusion-C} & 29  & 0.56 & 9.91 G & $10.1\times$ & \XSolidBrush & 4.39 & 191.33 & 81.09 & 52.31 \\
    \textbf{Flexiffusion-D} & 24  & 0.33 & 5.98 G & $16.7\times$ & \XSolidBrush & 6.71 & 174.33 & 77.96 & 50.71 \\
    \bottomrule
    \end{tabular}}
    \vspace{-5pt}
    \caption{Comparison on ImageNet $(256\times256)$. * denotes for using ImageNet $(64\times64)$. Speedup is calculated by the average MACs.}
    \label{tab: performance_ldm}
\end{table*}

\begin{table*}[t]
    \centering
    \resizebox{0.98\linewidth}{!}{
    \begin{tabular}{c||cccc|cccc|cccc}
    \toprule
    \multirow{2}*{\textbf{Method}} & \multicolumn{4}{c|}{\textbf{CIFAR-10} \bm{$32\times32$}} & \multicolumn{4}{c|}{\textbf{Bedroom} \bm{$256\times256$}} & \multicolumn{4}{c}{\textbf{Church} \bm{$256\times256$}} \\
    \cmidrule(r){2-13}
    ~ & \textbf{NFE} $\downarrow$ & \textbf{avg. MACs} $\downarrow$ & \textbf{Speedup} $\uparrow$ & \textbf{FID} $\downarrow$ & \textbf{NFE} $\downarrow$ & \textbf{avg. MACs} $\downarrow$ & \textbf{Speedup} $\uparrow$ & \textbf{FID} $\downarrow$ & \textbf{NFE} $\downarrow$ & \textbf{avg. MACs} $\downarrow$ & \textbf{Speedup} $\uparrow$ & \textbf{FID} $\downarrow$ \\
    \midrule
    DDIM  & 100 & 6.10 G & $1.0\times$ & 4.19 & 100 & 248.7 G & $1.0\times$ & 6.62 & 100 & 248.7 G & $1.0\times$ & 10.58 \\
    OMS-DPM  & 19 & 0.42 G & $14.5\times$ & 48.45 & - & - & - & - & - & - & - & - \\
    DDSM  & 1000 & 62.00 G & $0.1\times$ & 3.55 & - & - & - & - & - & - & - & - \\
    \midrule
    DeepCache-B & 100 & 3.01 G & $2.0\times$ & 5.82  & 100 & 156.0 G & $1.6\times$ & 9.49  & 100 & 156.0 G & $1.4\times$ & 13.78  \\
    DeepCache-C & 100 & 2.63 G & $2.3\times$ & 10.41 & 100 & 144.4 G & $1.6\times$ & 17.28 & 100 & 144.4 G & $1.7\times$ & 22.65  \\
    DeepCache-D & 100 & 2.42 G & $2.5\times$ & 17.90 & 100 & 138.7 G & $1.6\times$ & 38.84 & 100 & 138.7 G & $1.8\times$ & 37.51 \\
    \midrule
    \textbf{Flexiffusion-B} & 70 & 2.80 G & $2.2\times$ & 5.75 & 57 & 108.0 G & $2.2\times$ & 7.35 & 59 & 113.4 G & $2.2\times$ & 12.33 \\
    \textbf{Flexiffusion-C} & 65 & 2.50 G & $2.4\times$ & 6.58 & 53 & 99.3 G  & $2.5\times$ & 7.05 & 52 & 99.1 G  & $2.5\times$ & 12.63 \\
    \textbf{Flexiffusion-D} & 54 & 1.97 G & $3.1\times$ & 7.19 & 50 & 87.8 G  & $2.8\times$ & 9.01 & 44 & 84.9 G  & $2.9\times$ & 14.31 \\
    \bottomrule
    \end{tabular}}
    \vspace{-5pt}
    \caption{Image generation quality and computing cost on CIFAR-10, LSUN Bedroom and Church. The default number of time step is 100.}
    \label{tab: performance_ddpm}
\end{table*}

\begin{table*}[t]
    \centering
    \resizebox{0.98\linewidth}{!}{
    \begin{tabular}{c||ccccc|cccccc}
    \toprule
    \multirow{2}*{\textbf{Method}} & \multicolumn{5}{c|}{\textbf{Parti-Prompts} \bm{$512\times512$}} & \multicolumn{6}{c}{\textbf{MS-COCO \bm{$512\times512$}}} \\
    \cmidrule(r){2-12}
    ~ & \textbf{NFE} $\downarrow$ & \textbf{s/image} $\downarrow$ & \textbf{avg. MACs} $\downarrow$ & \textbf{Speedup} $\uparrow$ & \textbf{CLIP Score}  $\uparrow$ & \textbf{NFE} $\downarrow$ & \textbf{s/image} $\downarrow$ & \textbf{avg. MACs} $\downarrow$ & \textbf{Speedup} $\uparrow$ & \textbf{CLIP Score} $\uparrow$ & \textbf{FID} $\downarrow$  \\
    \midrule
    PLMS & 50 & 3.01 & 338.76 G & $1.0\times$ & 29.76 & 50 & 3.11 & 338.76 G & $1.0\times$ & 30.37 & 22.19 \\
    \midrule
    DeepCache-A & 50 & 2.06 & 198.03 G & $1.7\times$ & 29.80 & 50 & 2.18 & 198.03 G & $1.7\times$ & 30.42  & 22.19 \\
    DeepCache-B & 50 & 1.54 & 130.45 G & $2.6\times$ & 29.51 & 50 & 1.61 & 130.45 G & $2.6\times$ & 30.32 & 21.33 \\
    DeepCache-C & 50 & 1.35 &  85.54 G & $3.9\times$ & 29.02 & 50 & 1.61 & 85.54 G & $3.9\times$ &  29.65 & 21.64  \\
    \midrule
    \textbf{Flexiffusion-A} & 25 & 0.97 & 87.03 G & $3.9\times$ & 29.83 & 25 & 1.07 & 88.90 G & $3.8\times$ & 30.45 & 21.28\\
    \textbf{Flexiffusion-B} & 24 & 0.86 & 77.17 G & $4.3\times$ & 29.60 & 26 & 0.96 & 79.00 G & $4.5\times$ & 30.40 & 20.99 \\
    \textbf{Flexiffusion-C} & 15 & 0.76 & 68.26 G & $5.0\times$ & 29.37 & 16 & 0.86 & 66.32 G & $5.1\times$ & 30.11 & 21.27\\
    \bottomrule
    \end{tabular}}
    \vspace{-5pt}
    \caption{Comparison of text-to-image generation results of Stable Diffusion. The default number of time step is 50.}
    \label{tab: performance_sd}
\end{table*}

\section{Experiment}
\label{sec: experiment}
\subsection{Experiment Settings}
\label{subsec: experiment settings}
\textbf{Settings for Diffusion Models.} We evaluate our approach on three widely-used frameworks: DDPM \cite{ho2020denoising}, LDM \cite{rombach2022high}, and Stable Diffusion V1.5 \cite{rombach2022high}. We conduct experiments using DDIM \cite{song2020denoising}, PLMS \cite{liu2022pseudo}, DPM-Solver \cite{lu2022dpm} sampler. We consider six different datasets, including CIFAR10 \cite{krizhevsky2009learning}, LSUN Bedroom and Church \cite{yu2015lsun}, ImageNet \cite{deng2009imagenet}, Parti-Prompts \cite{yu2022scaling} and MS-COCO \cite{lin2014microsoft}. 


\noindent\textbf{Settings for NAS.}  We define three searchable dimensions as discussed in \cref{subsec:segment-wise search space}. During the search, we use MACs as the primary resource constraint, ensuring that each candidate is considered valid only if its MACs are below the specified maximum limit. The search space settings are varying across different diffusion framework.
For detailed information of the space and search settings, please refer to the supplementary material.

\noindent\textbf{Baselines.} We select DeepCache \cite{ma2023deepcache} as the primary baseline, which is the SOTA training-free acceleration method. In the following, we use DeepCache-A/B/C/D to represent models using uniform interval settings of $\{2,5,10,20\}$. Models with the symbol + denote quadratic cache variants, which reports better performance than the uniform ones. The \textit{branch} is 2/1/2 for DDPM, LDM and Stable Diffusion. We also compare with two latest NAS methods, OMS-DPM \cite{liu2023oms} and DDSM \cite{yang2023denoising}, which are training-based. The model performance of OMS-DPM is based on open-source models and DDSM is based on reports.

\noindent\textbf{Evaluation Metrics.} As for inference cost, we measure the number of function evaluations (NFE), sampling time (s/image), and average Multiply-Accumulate Operations (average MACs). NFE typically equals the number of time steps, except in Flexiffusion due to the \textit{null} step. Sampling time is measured on an NVIDIA RTX 3090, and the average MACs are calculated based on the time steps. As for model performance, we measure the FID \cite{heusel2017gans}, Inception Score (IS) \cite{salimans2016improved}, Precisions and Recall \cite{kynkaanniemi2019improved}. As for the text-to-image model, we measure the CLIP Score \cite{hessel2021clipscore}. The calculation of these metrics is following DeepCache.




\subsection{Comparison on LDM} 
\cref{tab: performance_ldm} shows the results on ImageNet. Our method uses LDM with the 250-step DDIM sampler as DeepCache. Our searched diffusion model, Flexiffusion-A/B/C/D, reports fewer resource consumptions, faster generation speed and higher image quality than the counterpart in DeepCache. While DeepCache-D+, achieve a $10.6\times$ acceleration, Flexiffusion-D attains a $16.7\times$ speedup for LDM with an improved FID of 6.71. Image examples are shown in supplementary material, and please refer to them.

\subsection{Comparison on DDPM} 
The results of are shown in \cref{tab: performance_ddpm}, including comparison on CIFAR-10, LSUN Bedroom and Church. In low-budget cases, while DeepCache-C/D are suffering from low image quality, Flexiffusion-C/D report significantly lower FID with lower latency.  Although the NAS-based methods OMS-DPM achieves faster speeds and DDSM attains a better FID, Flexiffusion offers a superior speed-quality trade-off, especially given its training-free nature. Please refer to the supplementary material for more visual comparisons. 

\subsection{Comparison on Stable Diffusion} 
Flexiffusion employs a 50-step PLMS sampler as DeepCache. \cref{tab: performance_sd} summarizes the model performance on Parti-Prompts and MS-COCO. Images generated by Flexiffusion demonstrate higher quality and improved alignment with textual prompts compared to DeepCache. Additionally, Flexiffusion achieves a maximum speedup of $5.1\times$ over the original Stable Diffusion with PLMS.

\subsection{Ablation Study for rFID}
\label{subsubsec: study for rfid}
We assess both the time cost and ranking consistency for various metrics, including FID-50k, FID-1k, MSE, and rFID. These metrics are evaluated on 1,000 random models from the search space on LDM/ImageNet using an NVIDIA RTX 3090. The computational overhead for FID and its variants is approximately 12.3 seconds per model, while MSE has a lower overhead of 6.6 seconds per model. As shown in \cref{tab:relative fid}, FID-50k achieves a Kendall’s $\tau$ of 1.00 but requires a significantly higher computational cost of 2867.9 GPU hours. In comparison, rFID and its variants demonstrate a favorable trade-off between efficiency and effectiveness. 
 
\begin{table}[h]
    \centering
    \resizebox{0.99\linewidth}{!}{
    \begin{tabular}{cccc}
    \toprule
    \multirow{2}*{\textbf{Metric}}  &  \multirow{2}*{\textbf{Num of Image} $\downarrow$} &   \textbf{Time Cost}  $\downarrow$ & \multirow{2}*{\textbf{Kendall's $\tau$} $\uparrow$}  \\
    ~ & ~ & (GPU Hours) & ~ \\
    \midrule
    FID-50k            & 50000 & 2867.9 & 1.00 \\
    FID-1k             & 1000  & 58.3 & 0.44 \\
    MSE                & 1000  & 57.9 & 0.31 \\
    \midrule
    rFID-2k      & 2000  & 117.0 & 0.80 \\
    rFID-1k      & 1000  & 58.3 & 0.76 \\
    rFID-0.2k      & 200  & 11.6 & 0.57 \\
    \bottomrule
    \end{tabular}}
    \caption{Cost and ranking consistency across various metrics.}
    \label{tab:relative fid}
\end{table}

\subsection{Search Cost Analysis}
\label{subsec: search cost analysis}

The majority cost of our proposed segment-wise search comes from the candidate evaluation, while the mutation cost is negligible.  In \cref{tab:search cost}, we report the search cost in different datasets and frameworks. We note that since the search process is training-free, which is not need for retraining the denoising U-Net, and faster due to our proposed rFID. Compared to DDSM \cite{yang2023denoising}, which requires 320 GPU hours on CIFAR-10, plus an additional 502 GPU hours for training, our Flexiffusion only require 3 hours. The low search cost allows Flexiffusion to search more candidates.

\begin{table}[ht]
    \centering
    \resizebox{\linewidth}{!}{
    \begin{tabular}{ccccc}
    \toprule
    \multirow{2}*{Dataset} & \multirow{2}*{Framework} & \multirow{2}*{Resolution} & Budget & Total Cost\\
    ~ & ~ & ~ & (MACs) & (GPU Hours) \\
    \midrule
    CIFAR      & DDPM & $32$  & 3 G   & 3 \\
    Bedroom    & DDPM & $256$ & 120 G & 116 \\
    Church     & DDPM & $256$ & 120 G & 116 \\
    ImageNet   & LDM  & $256$ & 40 G  & 134 \\
    Parti      & SD   & $512$ & 90 G  & 34 \\
    COCO       & SD   & $512$ & 90 G  & 66 \\
    \bottomrule
    \end{tabular}}
    \vspace{-5pt}
    \caption{Search cost of Flexiffusion measured by an NVIDIA RTX 3090. The Budget refers to the maximum average MACs for candidates in the search process.  }
    \label{tab:search cost}
\end{table}

\subsection{Combine with Other Acceleration Approaches}
We also integrate Flexiffusion with other training-free acceleration techniques to demonstrate its compatibility.

\vspace{1pt}
\noindent\textbf{High order sampler.}
We also combine Flexiffusion with DPM-Solver \cite{lu2022dpm}. To support multi-order in the generation, we set the \textit{interval} choices $\{1,2,3\}$ representing order-1/2/3 for each segment. We fix the \textit{N-segment} to 16 and \textit{branch} choices are $\{1,2,3,4,5,6,12\}$. DeepCache uses a fixed $interval=2$, $branch=1$, and the third-order DPM-Solver. As shown in \cref{tab:dpm-solver}, while OMS-DPM and DeepCache report low image quality, Flexiffusion reports significant speedup with lower FID compared to DPM-Solver-3.

\begin{table}[ht]
    \centering
    \resizebox{\linewidth}{!}{
    \begin{tabular}{c||ccccc}
    \toprule
    \textbf{Method} & \textbf{NFE} $\downarrow$ & \textbf{ms/image} $\downarrow$ & \textbf{MACs} $\downarrow$ & \textbf{Speedup} $\uparrow$ & \textbf{FID} $\downarrow$\\
    \midrule
    \multirow{2}*{DPM-Solver-3} & 15 & 11.50 & 90.97 G & $1.0\times$ & 4.73 \\
    ~                           & 10 & 8.74  & 60.65 G & $1.5\times$ & 6.40 \\
    \midrule
    \multirow{2}*{OMS-DPM} & 25 & 11.12 & 88.42 G & $1.0\times$ & 23.81 \\
    ~                      & 16 &  7.54 & 47.89 G & $1.9\times$ & 24.29 \\
    \midrule
    \multirow{2}*{DeepCache}  & 25 &  12.22 & 94.28 G & $1.0\times$ & 10.64 \\
    ~                         & 15 &   8.62 &  57.2 G & $1.6\times$ & 23.33 \\
    \midrule
    \multirow{3}*{\textbf{Flexiffusion}} & 12 &  9.78 & 70.87 G & $1.3\times$ & 4.52 \\
    ~                                    & 10 &  8.20 & 55.87 G & $1.7\times$ & 5.22 \\
    ~                                    & 8  &  7.12 & 46.41 G & $2.0\times$ & 6.39 \\
    
    \bottomrule
    \end{tabular}}
    \vspace{-5pt}
    \caption{Comparison of using DPM-Solver on DDPM/CIFAR-10.}
    \label{tab:dpm-solver}
\end{table}

\vspace{1pt}
\noindent\textbf{Plug-and-use module.} Flexiffusion is compatible with the widely-used module ToMe \cite{bolya2023tomesd}. As shown in \cref{tab:tome}, Flexiffusion demonstrates better trade-offs between generation speed and quality, both with and without ToMe. We also provide an experiment about using ToMe in the search process, please refer to the supplementary material. 

\begin{table}[h]
    \centering
    \resizebox{\linewidth}{!}{
    \begin{tabular}{c||cc|cc}
    \toprule
    \multirow{2}*{\textbf{Method}} & \multicolumn{2}{c|}{\textbf{w/o ToMe}} & \multicolumn{2}{c}{\textbf{w/ ToMe}} \\
    \cmidrule(r){2-5}
    ~ & \textbf{s/image} $\downarrow$ & \textbf{Clip Score}  $\uparrow$ & \textbf{s/image} $\downarrow$ & \textbf{Clip Score} $\uparrow$ \\
    \midrule
    PLMS & 3.01 & 29.76 & 2.19 & 29.77 \\
    \midrule
    DeepCache-A & 2.06 & 29.80 & 1.43 & 29.78 \\
    DeepCache-B & 1.54 & 29.51 & 0.97 & 29.50 \\
    DeepCache-C & 1.35 & 29.02 & 0.83 & 28.85 \\
    \midrule
    \textbf{Flexiffusion-A} & 0.97 & 29.82 & 0.67 & 29.78\\
    \textbf{Flexiffusion-B} & 0.86 & 29.68 & 0.60 & 29.63\\
    \textbf{Flexiffusion-C} & 0.76 & 29.40 & 0.53 & 29.41\\
    \bottomrule
    \end{tabular}}
    \vspace{-5pt}
    \caption{w/o and w/ ToMe comparison on Parti-Prompts.}
    \label{tab:tome}
\end{table}
\vspace{-1em}

\section{Conclusion}

In this paper, we propose Flexiffusion, a training-free NAS framework that speedup diffusion models by co-optimizing generation schedules and architectures via segment-wise search. By segmenting denoising schedule into equal-length units with adaptive computations, Flexiffusion minimizes generative redundancy while remaining high quality. Experiments show Flexiffusion surpasses existing training-free methods in speed-quality trade-offs, offering a practical path toward efficient high-fidelity diffusion models.

{
    \small
    \bibliographystyle{ieeenat_fullname}
    \bibliography{main}

\begin{thebibliography}{50}
\providecommand{\natexlab}[1]{#1}
\providecommand{\url}[1]{\texttt{#1}}
\expandafter\ifx\csname urlstyle\endcsname\relax
  \providecommand{\doi}[1]{doi: #1}\else
  \providecommand{\doi}{doi: \begingroup \urlstyle{rm}\Url}\fi

\bibitem[Bao et~al.(2022)Bao, Li, Zhu, and Zhang]{bao2022analytic}
Fan Bao, Chongxuan Li, Jun Zhu, and Bo Zhang.
\newblock Analytic-dpm: an analytic estimate of the optimal reverse variance in diffusion probabilistic models.
\newblock \emph{arXiv preprint arXiv:2201.06503}, 2022.

\bibitem[Bolya and Hoffman(2023)]{bolya2023tomesd}
Daniel Bolya and Judy Hoffman.
\newblock Token merging for fast stable diffusion.
\newblock \emph{CVPR Workshop on Efficient Deep Learning for Computer Vision}, 2023.

\bibitem[Choi et~al.(2021)Choi, Kim, Jeong, Gwon, and Yoon]{choi2021ilvr}
Jooyoung Choi, Sungwon Kim, Yonghyun Jeong, Youngjune Gwon, and Sungroh Yoon.
\newblock Ilvr: Conditioning method for denoising diffusion probabilistic models.
\newblock \emph{arXiv preprint arXiv:2108.02938}, 2021.

\bibitem[Choi et~al.(2022)Choi, Lee, Shin, Kim, Kim, and Yoon]{choi2022perception}
Jooyoung Choi, Jungbeom Lee, Chaehun Shin, Sungwon Kim, Hyunwoo Kim, and Sungroh Yoon.
\newblock Perception prioritized training of diffusion models.
\newblock In \emph{Proceedings of the IEEE/CVF Conference on Computer Vision and Pattern Recognition}, pages 11472--11481, 2022.

\bibitem[Deja et~al.(2022)Deja, Kuzina, Trzcinski, and Tomczak]{deja2022analyzing}
Kamil Deja, Anna Kuzina, Tomasz Trzcinski, and Jakub Tomczak.
\newblock On analyzing generative and denoising capabilities of diffusion-based deep generative models.
\newblock \emph{Advances in Neural Information Processing Systems}, 35:\penalty0 26218--26229, 2022.

\bibitem[Deng et~al.(2009)Deng, Dong, Socher, Li, Li, and Fei-Fei]{deng2009imagenet}
Jia Deng, Wei Dong, Richard Socher, Li-Jia Li, Kai Li, and Li Fei-Fei.
\newblock Imagenet: A large-scale hierarchical image database.
\newblock In \emph{2009 IEEE conference on computer vision and pattern recognition}, pages 248--255. Ieee, 2009.

\bibitem[Dhariwal and Nichol(2021)]{dhariwal2021diffusion}
Prafulla Dhariwal and Alexander Nichol.
\newblock Diffusion models beat gans on image synthesis.
\newblock \emph{Advances in neural information processing systems}, 34:\penalty0 8780--8794, 2021.

\bibitem[Fang et~al.(2024)Fang, Ma, and Wang]{fang2024structural}
Gongfan Fang, Xinyin Ma, and Xinchao Wang.
\newblock Structural pruning for diffusion models.
\newblock \emph{Advances in neural information processing systems}, 36, 2024.

\bibitem[Goodfellow et~al.(2014)Goodfellow, Pouget-Abadie, Mirza, Xu, Warde-Farley, Ozair, Courville, and Bengio]{goodfellow2014generative}
Ian Goodfellow, Jean Pouget-Abadie, Mehdi Mirza, Bing Xu, David Warde-Farley, Sherjil Ozair, Aaron Courville, and Yoshua Bengio.
\newblock Generative adversarial nets.
\newblock \emph{Advances in neural information processing systems}, 27, 2014.

\bibitem[He et~al.(2021)He, Zhao, and Chu]{he2021automl}
Xin He, Kaiyong Zhao, and Xiaowen Chu.
\newblock Automl: A survey of the state-of-the-art.
\newblock \emph{Knowledge-based systems}, 212:\penalty0 106622, 2021.

\bibitem[He et~al.(2024)He, Liu, Liu, Wu, Zhou, and Zhuang]{he2024ptqd}
Yefei He, Luping Liu, Jing Liu, Weijia Wu, Hong Zhou, and Bohan Zhuang.
\newblock Ptqd: Accurate post-training quantization for diffusion models.
\newblock \emph{Advances in Neural Information Processing Systems}, 36, 2024.

\bibitem[Hessel et~al.(2021)Hessel, Holtzman, Forbes, Bras, and Choi]{hessel2021clipscore}
Jack Hessel, Ari Holtzman, Maxwell Forbes, Ronan~Le Bras, and Yejin Choi.
\newblock Clipscore: A reference-free evaluation metric for image captioning.
\newblock \emph{arXiv preprint arXiv:2104.08718}, 2021.

\bibitem[Heusel et~al.(2017)Heusel, Ramsauer, Unterthiner, Nessler, and Hochreiter]{heusel2017gans}
Martin Heusel, Hubert Ramsauer, Thomas Unterthiner, Bernhard Nessler, and Sepp Hochreiter.
\newblock Gans trained by a two time-scale update rule converge to a local nash equilibrium.
\newblock \emph{Advances in neural information processing systems}, 30, 2017.

\bibitem[Hinton et~al.(2015)Hinton, Vinyals, and Dean]{hinton2015distilling}
Geoffrey Hinton, Oriol Vinyals, and Jeff Dean.
\newblock Distilling the knowledge in a neural network.
\newblock \emph{arXiv preprint arXiv:1503.02531}, 2015.

\bibitem[Ho et~al.(2020)Ho, Jain, and Abbeel]{ho2020denoising}
Jonathan Ho, Ajay Jain, and Pieter Abbeel.
\newblock Denoising diffusion probabilistic models.
\newblock \emph{Advances in neural information processing systems}, 33:\penalty0 6840--6851, 2020.

\bibitem[Kim et~al.(2023)Kim, Song, Castells, and Choi]{kim2023bksdm}
Bo-Kyeong Kim, Hyoung-Kyu Song, Thibault Castells, and Shinkook Choi.
\newblock Bk-sdm: A lightweight, fast, and cheap version of stable diffusion.
\newblock \emph{arXiv preprint arXiv:2305.15798}, 2023.

\bibitem[Kingma and Welling(2013)]{kingma2013auto}
Diederik~P Kingma and Max Welling.
\newblock Auto-encoding variational bayes.
\newblock \emph{arXiv preprint arXiv:1312.6114}, 2013.

\bibitem[Krizhevsky et~al.(2009)Krizhevsky, Hinton, et~al.]{krizhevsky2009learning}
Alex Krizhevsky, Geoffrey Hinton, et~al.
\newblock Learning multiple layers of features from tiny images.
\newblock 2009.

\bibitem[Kynk{\"a}{\"a}nniemi et~al.(2019)Kynk{\"a}{\"a}nniemi, Karras, Laine, Lehtinen, and Aila]{kynkaanniemi2019improved}
Tuomas Kynk{\"a}{\"a}nniemi, Tero Karras, Samuli Laine, Jaakko Lehtinen, and Timo Aila.
\newblock Improved precision and recall metric for assessing generative models.
\newblock \emph{Advances in neural information processing systems}, 32, 2019.

\bibitem[Li et~al.(2020)Li, Peng, Yuan, Wang, Liang, Lin, and Chang]{li2020block}
Changlin Li, Jiefeng Peng, Liuchun Yuan, Guangrun Wang, Xiaodan Liang, Liang Lin, and Xiaojun Chang.
\newblock Block-wisely supervised neural architecture search with knowledge distillation.
\newblock In \emph{Proceedings of the IEEE/CVF Conference on Computer Vision and Pattern Recognition}, pages 1989--1998, 2020.

\bibitem[Li et~al.(2021)Li, Tang, Wang, Peng, Wang, Liang, and Chang]{li2021bossnas}
Changlin Li, Tao Tang, Guangrun Wang, Jiefeng Peng, Bing Wang, Xiaodan Liang, and Xiaojun Chang.
\newblock Bossnas: Exploring hybrid cnn-transformers with block-wisely self-supervised neural architecture search.
\newblock In \emph{Proceedings of the IEEE/CVF International Conference on Computer Vision}, pages 12281--12291, 2021.

\bibitem[Li et~al.(2023)Li, Li, Zheng, Wu, Xiao, Wang, Zheng, Pan, Chao, and Ji]{li2023autodiffusion}
Lijiang Li, Huixia Li, Xiawu Zheng, Jie Wu, Xuefeng Xiao, Rui Wang, Min Zheng, Xin Pan, Fei Chao, and Rongrong Ji.
\newblock Autodiffusion: Training-free optimization of time steps and architectures for automated diffusion model acceleration.
\newblock In \emph{Proceedings of the IEEE/CVF International Conference on Computer Vision}, pages 7105--7114, 2023.

\bibitem[Li et~al.(2024)Li, Wang, Jin, Hu, Chemerys, Fu, Wang, Tulyakov, and Ren]{li2024snapfusion}
Yanyu Li, Huan Wang, Qing Jin, Ju Hu, Pavlo Chemerys, Yun Fu, Yanzhi Wang, Sergey Tulyakov, and Jian Ren.
\newblock Snapfusion: Text-to-image diffusion model on mobile devices within two seconds.
\newblock \emph{Advances in Neural Information Processing Systems}, 36, 2024.

\bibitem[Lin et~al.(2014)Lin, Maire, Belongie, Hays, Perona, Ramanan, Doll{\'a}r, and Zitnick]{lin2014microsoft}
Tsung-Yi Lin, Michael Maire, Serge Belongie, James Hays, Pietro Perona, Deva Ramanan, Piotr Doll{\'a}r, and C~Lawrence Zitnick.
\newblock Microsoft coco: Common objects in context.
\newblock In \emph{Computer Vision--ECCV 2014: 13th European Conference, Zurich, Switzerland, September 6-12, 2014, Proceedings, Part V 13}, pages 740--755. Springer, 2014.

\bibitem[Liu et~al.(2023)Liu, Ning, Lin, Yang, and Wang]{liu2023oms}
Enshu Liu, Xuefei Ning, Zinan Lin, Huazhong Yang, and Yu Wang.
\newblock Oms-dpm: Optimizing the model schedule for diffusion probabilistic models.
\newblock In \emph{International Conference on Machine Learning}, pages 21915--21936. PMLR, 2023.

\bibitem[Liu et~al.(2022)Liu, Ren, Lin, and Zhao]{liu2022pseudo}
Luping Liu, Yi Ren, Zhijie Lin, and Zhou Zhao.
\newblock Pseudo numerical methods for diffusion models on manifolds.
\newblock \emph{arXiv preprint arXiv:2202.09778}, 2022.

\bibitem[Lu et~al.(2022)Lu, Zhou, Bao, Chen, Li, and Zhu]{lu2022dpm}
Cheng Lu, Yuhao Zhou, Fan Bao, Jianfei Chen, Chongxuan Li, and Jun Zhu.
\newblock Dpm-solver: A fast ode solver for diffusion probabilistic model sampling in around 10 steps.
\newblock \emph{Advances in Neural Information Processing Systems}, 35:\penalty0 5775--5787, 2022.

\bibitem[Luhman and Luhman(2021)]{luhman2021knowledge}
Eric Luhman and Troy Luhman.
\newblock Knowledge distillation in iterative generative models for improved sampling speed.
\newblock \emph{arXiv preprint arXiv:2101.02388}, 2021.

\bibitem[Lyu et~al.(2022)Lyu, Xu, Yang, Lin, and Dai]{lyu2022accelerating}
Zhaoyang Lyu, Xudong Xu, Ceyuan Yang, Dahua Lin, and Bo Dai.
\newblock Accelerating diffusion models via early stop of the diffusion process.
\newblock \emph{arXiv preprint arXiv:2205.12524}, 2022.

\bibitem[Ma et~al.(2023)Ma, Fang, and Wang]{ma2023deepcache}
Xinyin Ma, Gongfan Fang, and Xinchao Wang.
\newblock Deepcache: Accelerating diffusion models for free.
\newblock \emph{arXiv preprint arXiv:2312.00858}, 2023.

\bibitem[Peebles and Xie(2023)]{peebles2023scalable}
William Peebles and Saining Xie.
\newblock Scalable diffusion models with transformers.
\newblock In \emph{Proceedings of the IEEE/CVF international conference on computer vision}, pages 4195--4205, 2023.

\bibitem[Real et~al.(2017)Real, Moore, Selle, Saxena, Suematsu, Tan, Le, and Kurakin]{real2017large}
Esteban Real, Sherry Moore, Andrew Selle, Saurabh Saxena, Yutaka~Leon Suematsu, Jie Tan, Quoc~V Le, and Alexey Kurakin.
\newblock Large-scale evolution of image classifiers.
\newblock In \emph{International conference on machine learning}, pages 2902--2911. PMLR, 2017.

\bibitem[Real et~al.(2019)Real, Aggarwal, Huang, and Le]{real2019regularized}
Esteban Real, Alok Aggarwal, Yanping Huang, and Quoc~V Le.
\newblock Regularized evolution for image classifier architecture search.
\newblock In \emph{Proceedings of the aaai conference on artificial intelligence}, pages 4780--4789, 2019.

\bibitem[Ren et~al.(2021)Ren, Xiao, Chang, Huang, Li, Chen, and Wang]{ren2021comprehensive}
Pengzhen Ren, Yun Xiao, Xiaojun Chang, Po-Yao Huang, Zhihui Li, Xiaojiang Chen, and Xin Wang.
\newblock A comprehensive survey of neural architecture search: Challenges and solutions.
\newblock \emph{ACM Computing Surveys (CSUR)}, 54\penalty0 (4):\penalty0 1--34, 2021.

\bibitem[Rombach et~al.(2022)Rombach, Blattmann, Lorenz, Esser, and Ommer]{rombach2022high}
Robin Rombach, Andreas Blattmann, Dominik Lorenz, Patrick Esser, and Bj{\"o}rn Ommer.
\newblock High-resolution image synthesis with latent diffusion models.
\newblock In \emph{Proceedings of the IEEE/CVF conference on computer vision and pattern recognition}, pages 10684--10695, 2022.

\bibitem[Ronneberger et~al.(2015)Ronneberger, Fischer, and Brox]{ronneberger2015u}
Olaf Ronneberger, Philipp Fischer, and Thomas Brox.
\newblock U-net: Convolutional networks for biomedical image segmentation.
\newblock In \emph{Medical image computing and computer-assisted intervention--MICCAI 2015: 18th international conference, Munich, Germany, October 5-9, 2015, proceedings, part III 18}, pages 234--241. Springer, 2015.

\bibitem[Salimans and Ho(2022)]{salimans2022progressive}
Tim Salimans and Jonathan Ho.
\newblock Progressive distillation for fast sampling of diffusion models.
\newblock \emph{arXiv preprint arXiv:2202.00512}, 2022.

\bibitem[Salimans et~al.(2016)Salimans, Goodfellow, Zaremba, Cheung, Radford, and Chen]{salimans2016improved}
Tim Salimans, Ian Goodfellow, Wojciech Zaremba, Vicki Cheung, Alec Radford, and Xi Chen.
\newblock Improved techniques for training gans.
\newblock \emph{Advances in neural information processing systems}, 29, 2016.

\bibitem[Shang et~al.(2023)Shang, Yuan, Xie, Wu, and Yan]{shang2023post}
Yuzhang Shang, Zhihang Yuan, Bin Xie, Bingzhe Wu, and Yan Yan.
\newblock Post-training quantization on diffusion models.
\newblock In \emph{Proceedings of the IEEE/CVF Conference on Computer Vision and Pattern Recognition}, pages 1972--1981, 2023.

\bibitem[Song et~al.(2020{\natexlab{a}})Song, Meng, and Ermon]{song2020denoising}
Jiaming Song, Chenlin Meng, and Stefano Ermon.
\newblock Denoising diffusion implicit models.
\newblock \emph{arXiv preprint arXiv:2010.02502}, 2020{\natexlab{a}}.

\bibitem[Song and Ermon(2019)]{song2019generative}
Yang Song and Stefano Ermon.
\newblock Generative modeling by estimating gradients of the data distribution.
\newblock \emph{Advances in neural information processing systems}, 32, 2019.

\bibitem[Song et~al.(2020{\natexlab{b}})Song, Sohl-Dickstein, Kingma, Kumar, Ermon, and Poole]{song2020score}
Yang Song, Jascha Sohl-Dickstein, Diederik~P Kingma, Abhishek Kumar, Stefano Ermon, and Ben Poole.
\newblock Score-based generative modeling through stochastic differential equations.
\newblock \emph{arXiv preprint arXiv:2011.13456}, 2020{\natexlab{b}}.

\bibitem[Song et~al.(2023)Song, Dhariwal, Chen, and Sutskever]{song2023consistencymodels}
Yang Song, Prafulla Dhariwal, Mark Chen, and Ilya Sutskever.
\newblock Consistency models, 2023.

\bibitem[Wimbauer et~al.(2024)Wimbauer, Wu, Schoenfeld, Dai, Hou, He, Sanakoyeu, Zhang, Tsai, Kohler, et~al.]{wimbauer2024cache}
Felix Wimbauer, Bichen Wu, Edgar Schoenfeld, Xiaoliang Dai, Ji Hou, Zijian He, Artsiom Sanakoyeu, Peizhao Zhang, Sam Tsai, Jonas Kohler, et~al.
\newblock Cache me if you can: Accelerating diffusion models through block caching.
\newblock In \emph{Proceedings of the IEEE/CVF Conference on Computer Vision and Pattern Recognition}, pages 6211--6220, 2024.

\bibitem[Yang et~al.(2023)Yang, Chen, Wang, Liu, and Chen]{yang2023denoising}
Shuai Yang, Yukang Chen, Luozhou Wang, Shu Liu, and Yingcong Chen.
\newblock Denoising diffusion step-aware models.
\newblock \emph{arXiv preprint arXiv:2310.03337}, 2023.

\bibitem[Yu et~al.(2015)Yu, Seff, Zhang, Song, Funkhouser, and Xiao]{yu2015lsun}
Fisher Yu, Ari Seff, Yinda Zhang, Shuran Song, Thomas Funkhouser, and Jianxiong Xiao.
\newblock Lsun: Construction of a large-scale image dataset using deep learning with humans in the loop.
\newblock \emph{arXiv preprint arXiv:1506.03365}, 2015.

\bibitem[Yu and Huang(2019)]{yu2019autoslim}
Jiahui Yu and Thomas Huang.
\newblock Autoslim: Towards one-shot architecture search for channel numbers.
\newblock \emph{arXiv preprint arXiv:1903.11728}, 2019.

\bibitem[Yu et~al.(2022)Yu, Xu, Koh, Luong, Baid, Wang, Vasudevan, Ku, Yang, Ayan, et~al.]{yu2022scaling}
Jiahui Yu, Yuanzhong Xu, Jing~Yu Koh, Thang Luong, Gunjan Baid, Zirui Wang, Vijay Vasudevan, Alexander Ku, Yinfei Yang, Burcu~Karagol Ayan, et~al.
\newblock Scaling autoregressive models for content-rich text-to-image generation.
\newblock \emph{arXiv preprint arXiv:2206.10789}, 2\penalty0 (3):\penalty0 5, 2022.

\bibitem[Zoph and Le(2016)]{zoph2016neural}
Barret Zoph and Quoc~V Le.
\newblock Neural architecture search with reinforcement learning.
\newblock \emph{arXiv preprint arXiv:1611.01578}, 2016.

\bibitem[Zoph et~al.(2018)Zoph, Vasudevan, Shlens, and Le]{zoph2018learning}
Barret Zoph, Vijay Vasudevan, Jonathon Shlens, and Quoc~V Le.
\newblock Learning transferable architectures for scalable image recognition.
\newblock In \emph{Proceedings of the IEEE conference on computer vision and pattern recognition}, pages 8697--8710, 2018.

\end{thebibliography}
}

\clearpage
\setcounter{page}{1}
\maketitlesupplementary

\renewcommand\thesection{\Alph{section}}
\setcounter{section}{0}
\setcounter{figure}{0}
\numberwithin{equation}{section}
\numberwithin{figure}{section}

\section{The Evolutionary Search}

\cref{alg:search} and \cref{alg:mutation} are the pseudo algorithms for the search algorithm and the mutation process in Flexiffusion. 


\begin{algorithm}[h]
\setstretch{1.1}
\caption{Segment-wise evolutionary search}
\label[algorithm]{alg:search}
\KwIn{rFID estimator $F(\cdot)$; mutation func $M(\cdot)$; cost estimator $C(\cdot)$; resource limitation $R$; max number of parents $n_p$ and children $n_c$; max size of population $n_{u}$} 
Initialize a population $\mathcal{P}$ with candidate $x$\\
\While{not reach the max iteration}{
    Randomly sample $\{x_1,\cdots\!,x_{n_p}\}$ from $\mathcal{P}$ \\
    Get the top-1 $x^*$ from $\{x_1,\cdots\!,x_{n_p}\}$ via $F(\cdot)$\\
    Create an empty children set $\mathcal{P}_c$ \\
    \While{$|\mathcal{P}_c| < n_c$ and not reach the max times}{
        $x_{new} \leftarrow M(x^*)$ \\
        \If{$C(x_{new})<R$}{
            Add $x_{new}$ to $\mathcal{P}_c$ \\
        }
    }
    $\mathcal{P} \leftarrow \mathcal{P} \cup \mathcal{P}_c$ and keep top-$n_{u}$ candidates in $\mathcal{P}$\\
    
}
\KwOut{$\mathcal{P}$}
\end{algorithm}

\begin{algorithm}[ht]
\setstretch{1.1}
\caption{Segment-wise mutation}
\label[algorithm]{alg:mutation}
\KwIn{parent $x$; number of mutation segments $n_m$; \textit{branch} choices $\mathcal{B}$; \textit{interval} choices $\mathcal{I}$; \textit{N-segment} choices $\mathcal{S}$} 
Randomly sample segments $\{x_1,\cdots\!,x_{n_m}\}$ from $x$\\
\For{$x_i\leftarrow x_1,\cdots\!,x_{n_m}$}{
Randomly choose a mutation mode from "Heavier" or "Lighter"\\
Randomly choose an dimension $\mathcal{D}\leftarrow\{\mathcal{B},\mathcal{I},\mathcal{S}\}$\\
\uIf{"Heavier"}{
$x_i \leftarrow $ the larger setting of $\mathcal{D}(x_i)$\\
} 
\ElseIf{"Lighter"} {
$x_i \leftarrow $ the smaller setting of $\mathcal{D}(x_i)$\\
}
}
\KwOut{Mutated $x$}
\end{algorithm}

\section{NAS Settings for Experiment}
\label{apd: nas settings for experments}

\subsection{Hyper-parameters of the Search Space}

\cref{tab: NAS experiment settings} illustrates the space setting for Flexiffuion in different frameworks and datasets under given resource budgets. 

\begin{table}[ht]
    \centering
    \caption{NAS settings for different frameworks and datasets}
    \resizebox{\linewidth}{!}{
    \begin{tabular}{cccccc}
    \toprule
    \multirow{2}*{Framework} & \multirow{2}*{Dataset} & Budgets & \multirow{2}*{\textit{N-segment}} & \multirow{2}*{\textit{branch}} & \multirow{2}*{\textit{interval}} \\
    ~ & ~ & (MACs) & ~ & ~ & ~ \\
    \midrule
    \multirow{9}*{DDPM} & \multirow{3}*{CIFAR-10} & 3.0 G & 19,20,21 & 1,2,3 & 1,2,3 \\
    ~ & ~ & 2.5 G & 19,20,21 & 1,2,3 & 1,2,3 \\
    ~ & ~ & 2.0 G & 19,20,21 & 1,2,3 & 1,2,3 \\
    \cmidrule(r){2-6}
    ~ & \multirow{3}*{Bedroom} & 110 G & 27,28,29 & 1,3,6 & 2,3 \\
    ~ & ~ & 100 G & 21,22,23 & 1,3,6 & 2,3 \\
    ~ & ~ & 90 G & 19,21,22 & 1,3,6 & 2,3 \\
    \cmidrule(r){2-6}
    ~ & \multirow{3}*{Church} & 110 G & 27,28,29 & 1,3,6 & 2,3 \\
    ~ & ~ & 100 G & 21,22,23 & 1,3,6 & 2,3 \\
    ~ & ~ & 90 G & 19,21,22 & 1,3,6 & 2,3 \\
    \midrule
    \multirow{4}*{LDM} & \multirow{4}*{ImageNet} & 40 G & 59,60,61 & 1,3,6 & 2,3 \\
    ~ & ~ & 16 G & 29,30,31 & 1,3,6 & 2,3 \\
    ~ & ~ & 10 G & 14,15,16 & 1,3,6 & 2,3 \\
    ~ & ~ & 6 G & 9,10,11 & 1,3,6 & 2,3 \\
    \midrule
    \multirow{6}*{SD} & \multirow{3}*{Parti-Prompts} & 90 G & 9,10,11 & 1,3,6 & 2,3 \\
    ~ & ~ & 80 G & 8,9,10 & 1,3,6 & 2,3 \\
    ~ & ~ & 70 G & 7,8,9 & 1,3,6 & 2,3 \\
    \cmidrule(r){2-6}
    ~ & \multirow{3}*{MS-COCO} & 90 G & 9,10,11 & 1,3,6 & 2,3 \\
    ~ & ~ & 80 G & 8,9,10 & 1,3,6 & 2,3 \\
    ~ & ~ & 70 G & 7,8,9 & 1,3,6 & 2,3 \\
    \bottomrule
    \end{tabular}}
    \label{tab: NAS experiment settings}
\end{table}

\subsection{Settings of the Segment-wise Search}
Our proposed search process in Flexiffusion is following \cref{alg:search}.  Specifically, the maximum number of search iterations is set to be $100$. The maximum number of parents $n_p=25$. The maximum number of children $n_c=5$. The maximum size of population $n_u$ is 300. The maximum number of children generation times is 200.  

As for the candidate evaluation using rFID estimator $F$, we set the number of generative images in $F$ to 1000 in CIFAR-10, Church, Bedroom and ImageNet dataset. Since the datasets are relatively small, the image number is set to be 250 and 500 for Parti-prompts and MS-COCO, respectively.

\begin{figure*}[t]
    \centering
    \includegraphics[width=0.99\linewidth]{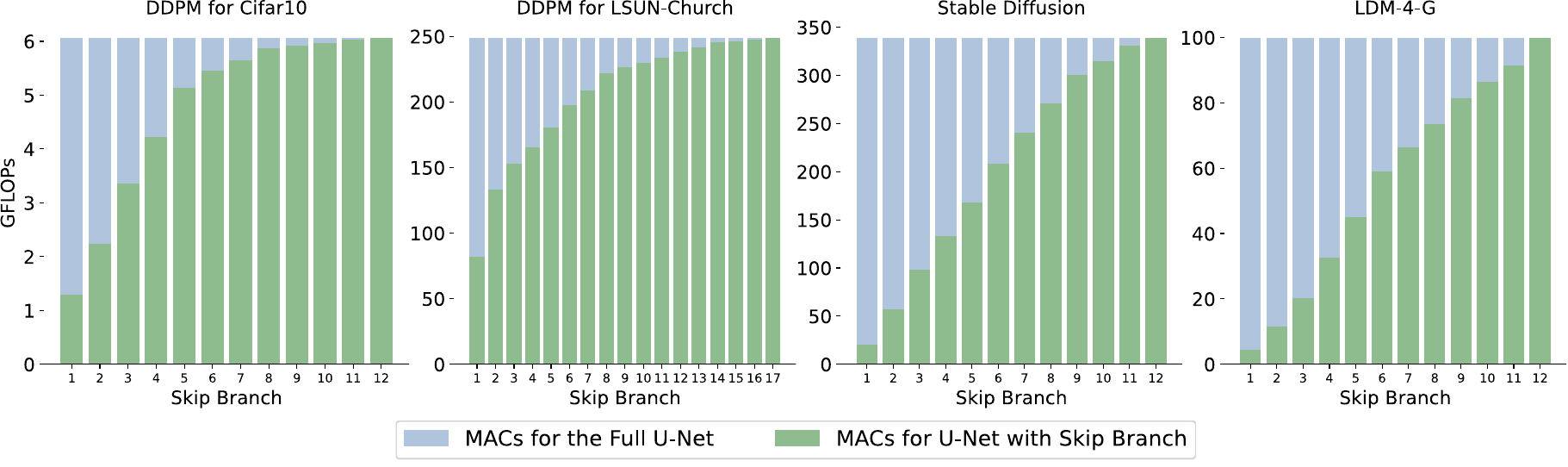}
    \caption{MACs for U-Net with different skip branch}
    \label{fig: flops}
\end{figure*}

\section{MACs for U-Net with Different Skip Branch}
\label{apd: macs for skip branch}
\cref{fig: flops} shows the MACs for U-Net in different frameworks with different skipping branches. We note that the increasing trends for MACs are different in different frameworks. For example, the MACs for $branch=6$ in DDPM on Cifar 10 is about $80\%$ of the whole U-Net, while the ratio in LDM-4-G is about $60\%$. These differences will affect the \textit{branch} choices for searching in different framework.

\section{Effect of Three Searchable Dimensions}

\begin{figure}[h]
\centering
\includegraphics[width=0.99\linewidth]{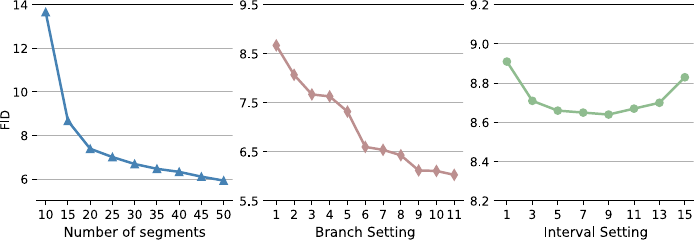}
\caption{The results of different dimension settings on LDM.}
\label{fig:dimension_effect}
\end{figure}

We study the impact of \textit{N-segment}, \textit{branch} and \textit{interval} on LDM/ImageNet. The default settings are $\{\textit{N-segment}, \textit{branch}, \textit{interval}\}=\{15, 5, 5\}$. The segments are set uniformly as DeepCache did. We only adjust one dimension at a time, and the results are illustrated in \cref{fig:dimension_effect}. It's obvious that the \textit{N-segment} is the most important dimension, which is highly correlated to FID, while the \textit{interval} is the less important one.

\begin{figure*}[t]
    \centering
    \includegraphics[width=0.99\linewidth]{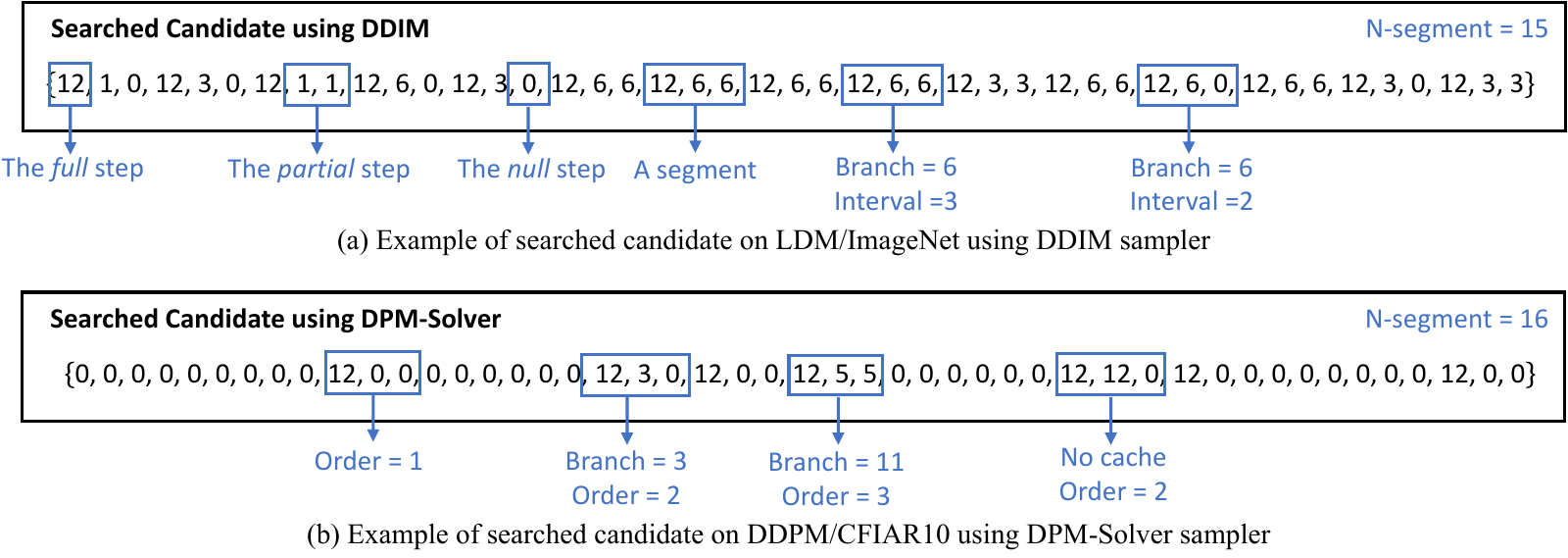}
    \caption{Example of searched diffusion models using DDIM or DPM-Solver.}
    \label{fig:searched schedule}
\end{figure*}

\section{Example of Searched Candidates}
\subsection{Searched Candidate using DDIM}
Considering a search space with settings: $\textit{N-segment} = \{15,16,17\}$, $\textit{branch} = \{1,3,6\}$, $\textit{interval} = \{2,3\}$, which is based on LDM-4-G using the DDIM sampler. The length of each segment is equal to the maximum \textit{interval}. The maximum branch in this framework is $12$. Therefore, we have roughly $(2\times3)^{15}+(2\times3)^{16}+(2\times3)^{17}\approx2\times10^{13}$ different candidates in the search space based on these settings. For simplification, we set each segment to start with a \textit{full} step (e.g., $12$), followed by several \textit{partial} steps, and ended by \textit{null} steps. Therefore, a segment can be defined by one \textit{branch} setting and one \textit{interval} setting. Accompany with a \textit{N-segment}, we can define a candidate diffusion model. We provide an example of a searched schedule as shown in \cref{fig:searched schedule} (a).

\subsection{Searched Candidate using DPM-Solver}
When it comes to DPM-Solver, we slightly adjust the rules of the search space. DPM-Solver is a type of high-order sampler for diffusion models in the continuous time domain, which naturally groups several adjacent model inference steps as an entity. For instance, DPM-Solver-3, the third-order solver, conducts three model inferences within a time interval.

In this case, the \textit{full}, \textit{partial} and \textit{null} step here is associated with the model inference (e.g., the NFE). To support multi-order computing in a candidate, we use the $\textit{interval} = \{1,2,3\}$ to represent the first, the second and the third order sampling. Considering a search space with settings:  $\textit{N-segment} = \{18\}$, $\textit{branch} = \{1, 2, 3, 4, 5, 6, 12\}$, $\textit{interval} = \{1,2,3\}$, which is based on CIFAR-10 using the DPM-Solver sampler. The maximum branch in this framework is $12$. Since we enable the branch choices for $\textit{branch} = 12$, there will be more than one \textit{full} steps in a segment. We provide an example of a searched schedule as shown in \cref{fig:searched schedule} (b).

\section{Number of Images for rFID}

Here, we randomly sample 1000 candidate diffusion models on LDM-4-G and measure the ranking correlation of rFID (using half precision) across various numbers of generated images by Kendall's $\tau$. \cref{tab: rfid with different number of imgs} reports that the ranking consistency decreases as the number of generated images decreases. The Time Cost denotes the total evaluation cost for all 1000 candidates on a single NVIDIA RTX 3090 GPU. Therefore, we recommend setting at least 1000 generated images for rFID to achieve a good trade-off between evaluation efficiency and accuracy.

\begin{table}[h]
    \centering
    \resizebox{\linewidth}{!}{
        \begin{tabular}{cccc}
        \toprule
        Num of Images & Time Cost (GPU Hours) & Kendall's $\tau$ \\
        \midrule
        5000 & 153.5 & 0.76 \\
        2000 & 62.4 & 0.75 \\
        1000 & 30.2 & 0.71 \\
        200 & 7.1 & 0.58 \\
        100 & 3.4 & 0.21 \\
        \bottomrule
        \end{tabular}}
    \caption{The effect of different numbers of generated images for calculating rFID}
    \label{tab: rfid with different number of imgs}
\end{table}


\section{Effectiveness of Flexiffusion}
We compare our searched diffusion model \cref{fig:searched schedule} (a) with a handcrafted schedule and a random schedule. The handcrafted schedule is constructed by 15 repeated segments as $\{12,6,6\}$. The "avg MACs" denote the average MACs of 100 steps in DDIM. \cref{tab: schedule effectiveness} reports the FID and speedup of three schedules. The searched schedule from Flexiffusion shows lower FID and lower computing cost, which indicates better speed and quality trade-offs.

\begin{table}[h]
    \centering
    \caption{FID comparison on LDM-4-G in ImageNet.}
    \begin{tabular}{c|ccc}
    \toprule
    \textbf{Schedule} & \textbf{avg MACs} & \textbf{FID-50K} & \textbf{Speedup} \\
    \midrule
    Handcrafted & 13.07G & 4.42 & $1.00\times$ \\
    Random & 11.14G & 5.48 & $1.17\times$ \\
    Flexiffusion & 9.91G & 4.39 & $1.32\times$ \\
    \bottomrule
    \end{tabular}
    \label{tab: schedule effectiveness}
\end{table}

\section{Discussion of Cache Mechanism on SD}
\label{apd: discussion}
\begin{figure}[h]
    \centering
    \includegraphics[width=0.99\linewidth]{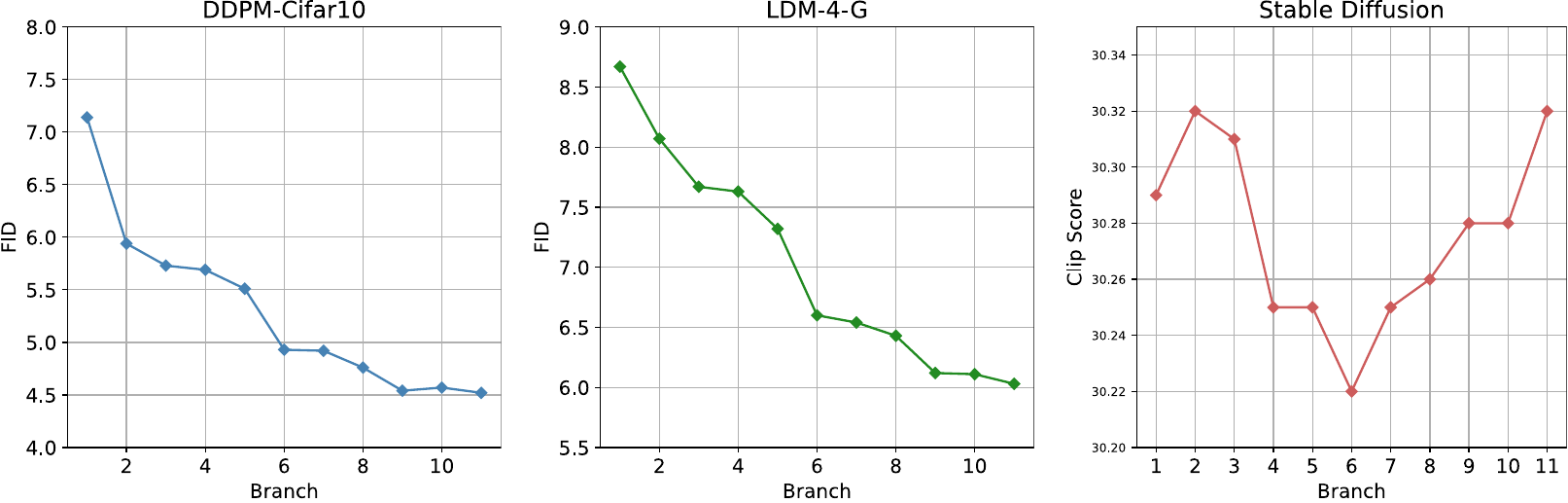}
    \caption{Effect of different skip branches on DDPM, LDM and Stable Diffusion}
    \label{fig: branches}
\end{figure}

During the experiment on Stable Diffusion using \textit{cache mechanism}, we observe an unstable performance decrease phenomenon. Unlike the gradual improvement trend of image quality in DDPM and LDM-4-G, the CLIP Score of different branch settings in Stable Diffusion shows numerical fluctuations, as shown in \cref{fig: branches}. This phenomenon is counterintuitive since a larger branch setting indicates more network blocks are involved in computing, which should positively affect model performance image quality, as in DDPM and LDM-4-G. More work needs to be done in the future to analyze this phenomenon.

Since this phenomenon is related to cache settings, our Flexiffusion can alleviate it via automatic searching for \textit{elastic branch} and report a better performance compared to handcrafted settings, as shown in \cref{tab: performance_sd}.

\section{Search Acceleration using ToMe}

We note that using ToMe for inference acceleration can further reduce the search cost. The table below reports a comparison of search costs on the Parti-Prompts dataset.  Searching incorporated with ToMe can obtain similar results with less time consumption. 

\begin{table}[h]
    \centering
    \resizebox{\linewidth}{!}{
    \begin{tabular}{c||cc|cc}
    \toprule
    \multirow{2}*{\textbf{Iteration}} & \multicolumn{2}{c|}{\textbf{w/o ToMe}} & \multicolumn{2}{c}{\textbf{w/ ToMe}} \\
    \cmidrule(r){2-5}
    ~ & \textbf{Time Cost} $\downarrow$ & \textbf{Clip Score}  $\uparrow$ & \textbf{Time Cost} $\downarrow$ & \textbf{Clip Score} $\uparrow$ \\
    \midrule
    10 & 3.21h & 29.41 & 2.30h & 29.35\\
    25 & 6.39h & 29.66 & 4.54h & 29.70\\
    50 & 13.46h & 29.82 & 9.38h & 29.80\\
    \bottomrule
    \end{tabular}}
\end{table}

\section{Generation Images}

\subsection{Prompts for Stable Diffusion}
Prompts for the teaser images:
\begin{itemize}
    \item "Sydney Opera House at sunset."
    \item "A photo of an astronaut in the space."
    \item "A lovely koala."
    \item "Unicorn galloping with rainbows."
    \item "Hot air balloons race over a meadow."
    \item "Delicious breakfast with vegan food on dishes."
\end{itemize}

\subsection{More Image Examples}
\label{apd: image examples}
\begin{figure*}[t]
    \flushright
    \includegraphics[width=0.95\linewidth]{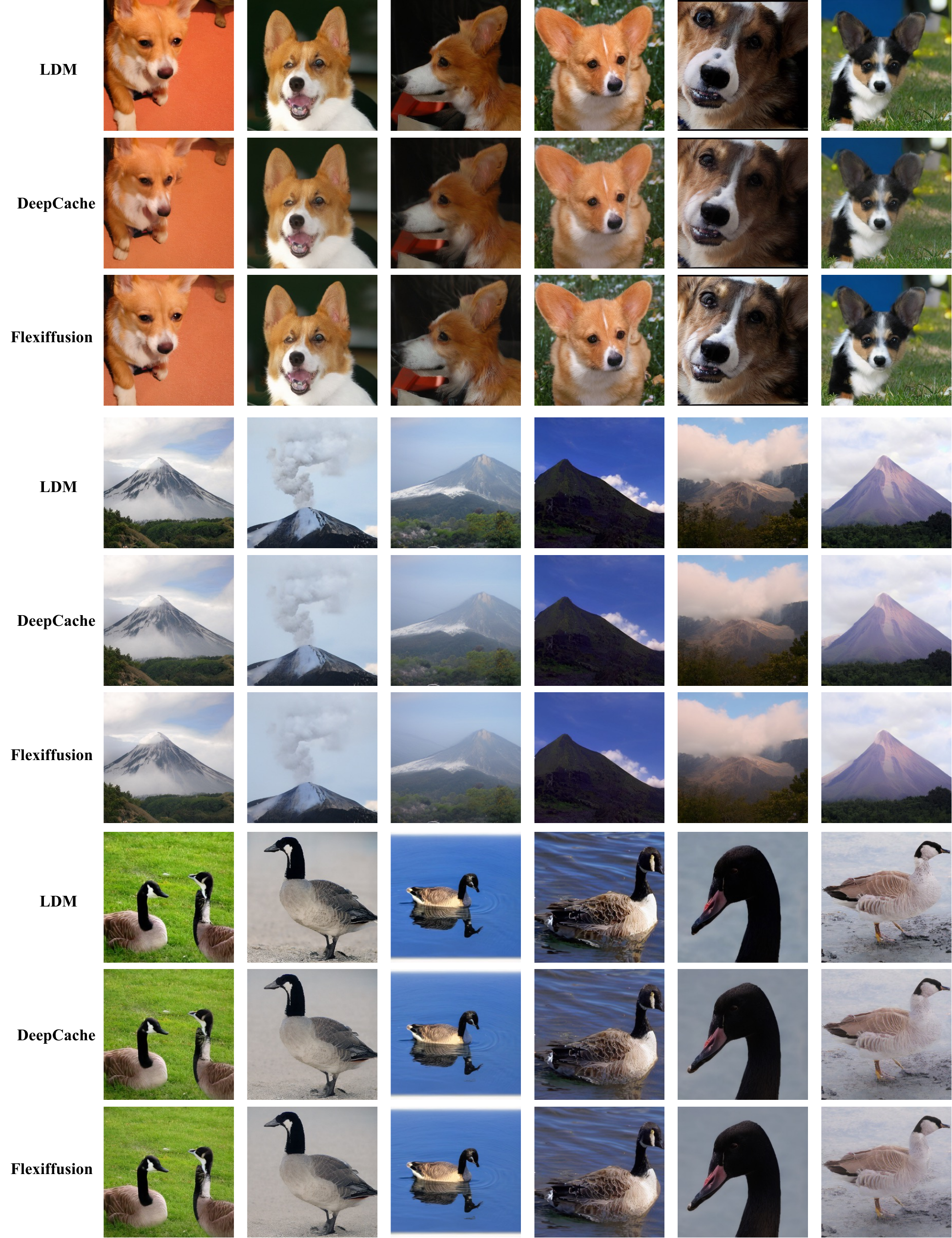}
    \caption{Comparisons of generative images from LDM, DeepCache-C+ and Flexiffusion-C on ImageNet. Flexiffusion is about $10.1\times$ faster than LDM and $1.4\times$ faster than DeepCache, while maintaining comparable image quality.}
    \label{fig: imagenet_examples}
\end{figure*}

\begin{figure*}[t]
    \flushright
    \includegraphics[width=0.95\linewidth]{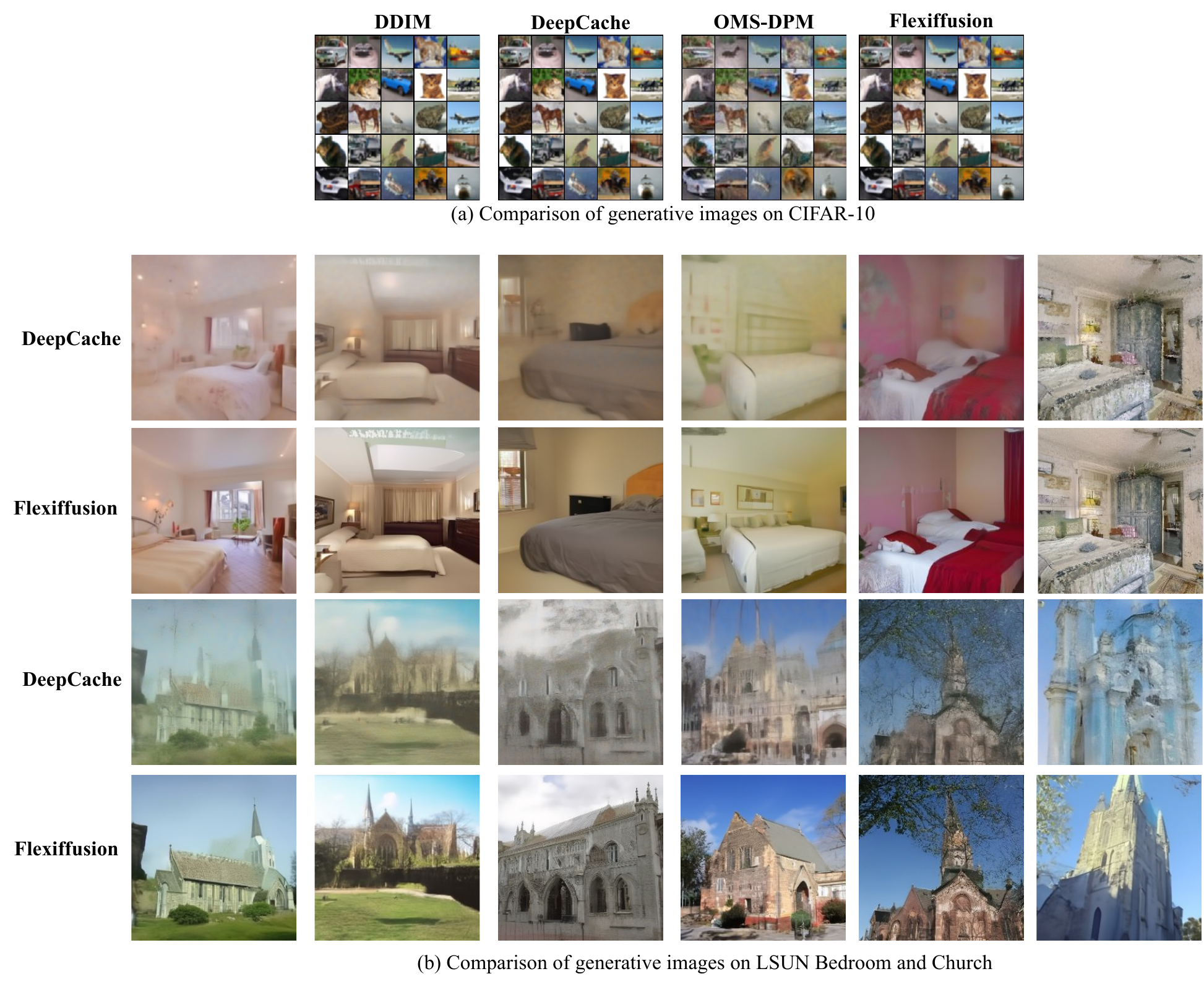}
    \caption{\textbf{(a):}Comparisons of generative images from DDIM, OMS-DPM, DeepCache-B and Flexiffusion-C on CIFAR-10. \textbf{(b):}Comparisons of generative images from DDIM, DeepCache-D and Flexiffusion-C on Bedroom and Church. Flexiffusion reports better balance between generative speed and quality.}
    \label{fig: other_examples}
\end{figure*}

We illustrated more visual comparisons in \cref{fig: imagenet_examples} and \cref{fig: other_examples}.

\end{document}